\documentclass[11pt]{article}

\usepackage[final]{acl}

\usepackage{times}
\usepackage{latexsym}
\usepackage[T1]{fontenc}

\usepackage[utf8]{inputenc}

\usepackage{microtype}

\usepackage{inconsolata}

\usepackage{amsmath}
\usepackage{amsfonts}
\usepackage{booktabs}
\usepackage{multirow}
\usepackage{graphicx}

\title{KLOD: Locality-Preserving Knowledge Editing via Non-Target Distribution Preservation}

\author{
  \textbf{Hojun Jeong\thanks{Equal contribution.}} \quad
  \textbf{Gyunyeop Kim\footnotemark[1]} \quad
  \textbf{Sangwoo Kang\thanks{Corresponding author.}}
\\
  Department of Computing, Gachon University
\\
  \texttt{\{wjdghwns12, gyop817, swkang\}@gachon.ac.kr}
}

\begin{document}
\maketitle
\begin{abstract}
Fine-tuning-based knowledge editing is simple and architecture-agnostic, but standard cross-entropy increases the edited target probability without explicitly constraining changes in the non-target output distribution. In sequential editing, such unconstrained redistribution can accumulate as distributional drift and contribute to locality degradation. We propose KLOD, a bounded and distribution-preserving objective for fine-tuning-based knowledge editing that separates the intended target update from distributions that should remain stable. KLOD stops target amplification once a probability threshold is reached, while preserving the target-excluded non-target distribution at target positions and the full next-token distribution at prefix positions. Experiments on CounterFact and ZsRE with Llama3-8B-Instruct and Qwen2.5-7B-Instruct show that KLOD substantially mitigates locality degradation while maintaining high edit reliability. The target probability threshold further provides a controllable Generalization--Locality trade-off. Ablation, multi-seed, and distributional KL analyses support the interpretation that KLOD's locality gains are associated with preserving output distributions rather than simply weakening the edit. Code is available on GitHub\footnote{\url{https://github.com/Hostoday/KLOD}}.
\end{abstract}

\section{Introduction}

Large language models (LLMs) store substantial factual knowledge in their parameters through large-scale pre-training, but this knowledge can become outdated or incorrect over time~\citep{DeCao2021}. Knowledge editing (KE) addresses this issue by modifying targeted facts while preserving unrelated behavior, which is especially important in sequential or lifelong editing settings where edits accumulate and may degrade locality, retention, and general capability~\citep{ContinualEditing,EditingAtScale}.

Prior KE methods largely follow a locate-and-edit paradigm, identifying internal computations or parameter regions associated with a factual association and modifying them directly~\citep{ROME,AlphaEdit}. Fine-tuning-based editing, by contrast, is architecture-agnostic and simple, but has often been treated as a disruptive baseline because gradient updates can affect broader model behavior~\citep{StandardFT}. Recent studies, however, show that its reliability and locality depend strongly on the objective, editable parameter selection, and training protocol~\citep{StandardFT,LocFT-BF}. This suggests that fine-tuning-based KE should be reconsidered as an objective design problem for locality-preserving editing. The resulting question is therefore not simply whether gradient-based fine-tuning can edit knowledge, but what objective should govern the update when edit success and preservation must be achieved simultaneously.

Standard fine-tuning typically maximizes the likelihood of the target answer under the rewrite prompt. However, KE requires more than target answer generation: the edited model should reflect the desired fact while remaining close to the pre-edit model on unrelated behavior. This is challenging in large-scale or sequential editing, where model editing can induce forgetting, locality degradation, and broader capability loss~\citep{ModelEditingHarms,EditingAtScale}. Therefore, a KE objective should control not only target likelihood but also the post-edit next-token distribution.

We identify an objective-level mismatch between standard softmax cross-entropy and locality-preserving editing. Cross-entropy increases target token probability, but due to softmax normalization this can be achieved either by raising the target logit or by lowering non-target logit mass. Thus, while effective for target learning, cross-entropy does not explicitly preserve the pre-edit relative distribution among non-target tokens. Complementary to prior observations that naive target maximization causes over-editing or token-level overfitting~\citep{OVERTONE,EditingOverfit}, we argue that non-target redistribution can accumulate as distributional drift in sequential fine-tuning-based editing, leading to locality degradation.

To address this issue, we propose KLOD, a bounded and distribution-preserving objective that decouples target editing from distribution preservation. At target prediction positions, KLOD keeps the renormalized non-target distribution, defined over all vocabulary tokens except the target token, close to that of the pre-edit model. This allows the intended target probability increase while preserving the structure of the non-target distribution. KLOD also uses a one-sided hinge objective based on target logit-odds, providing an edit signal only until the target probability reaches a threshold $\alpha$. At prefix positions, it preserves the full next-token distribution through KL regularization to suppress prompt-level drift.

Our contributions are threefold. First, we analyze how standard cross-entropy in fine-tuning-based KE fails to control non-target distribution changes. Second, we introduce KLOD, which combines bounded target optimization with non-target and prefix distribution preservation. Third, we show under established sequential KE protocols that KLOD mitigates the locality collapse of strong fine-tuning-based baselines while maintaining high reliability and providing a controllable Generalization--Locality operating curve. Seed-based experiments confirm that KLOD's locality advantage and the lower-Generalization/higher-Locality profile of the primary operating point are stable, while distributional KL analysis and ablations support the interpretation that the gains are associated with suppressing non-target distributional drift rather than merely weakening the edit. A broader discussion of related work is provided in Appendix~\ref{app:related_work}.

\section{Method}
\label{sec:method}

\subsection{Problem Setup}
\label{sec:problem_setup}

We consider knowledge editing for autoregressive language models. Let $f_{\theta_0}$ be the pre-edit model and $f_{\theta}$ the edited model. Each edit request consists of a rewrite prompt $x$ and a target sequence $y=(y_1,\ldots,y_m)$. The goal is to make the model generate $y$ given $x$, while preserving behavior unrelated to the edit.

Let
\begin{equation}
    s = x \oplus y .
\end{equation}
At position $t$, the model predicts the next token conditioned on $s_{\leq t}$, with logits
\begin{equation}
    z_t^\theta \in \mathbb{R}^{|\mathcal{V}|}
\end{equation}
and next-token distribution
\begin{equation}
    p_\theta(\cdot \mid s_{\leq t}) = \mathrm{softmax}(z_t^\theta).
\end{equation}
We denote the gold next token by $a_t=s_{t+1}$.

We define the target prediction positions as
\begin{equation}
    \mathcal{T}
    =
    \{t : s_{t+1} \text{ belongs to the target segment } y\}.
\end{equation}
The remaining valid next-token prediction positions are
\begin{equation}
    \mathcal{P}
    =
    \{t : t \text{ is a valid prediction position},\; t\notin\mathcal{T}\}.
\end{equation}
We refer to $\mathcal{P}$ as prefix positions, i.e., valid positions in the rewrite sequence that do not directly predict target tokens. Knowledge editing aims to increase target token probabilities at $\mathcal{T}$ while preventing unnecessary drift in the non-target distribution at $\mathcal{T}$ and the full next-token distribution at $\mathcal{P}$.

\subsection{Revisiting Cross-Entropy Fine-Tuning for Knowledge Editing}
\label{sec:ce_revisit}

Standard fine-tuning minimizes cross-entropy at target prediction positions. The target token probability is
\begin{equation}
    p_\theta(a_t\mid s_{\leq t})
    =
    \frac{
    \exp z_{t,a_t}^{\theta}
    }{
    \sum_{v\in\mathcal{V}}\exp z_{t,v}^{\theta}
    }.
    \label{eq:softmax_target}
\end{equation}
Thus,
\begin{equation}
    L_{\mathrm{CE}}
    =
    -\frac{1}{|\mathcal{T}|}
    \sum_{t\in\mathcal{T}}
    \log p_\theta(a_t\mid s_{\leq t}).
    \label{eq:ce}
\end{equation}
For a single position,
\begin{equation}
    -\log p_\theta(a_t\mid s_{\leq t})
    =
    -z_{t,a_t}^{\theta}
    +
    \log\sum_{v\in\mathcal{V}}\exp z_{t,v}^{\theta}.
    \label{eq:ce_logit}
\end{equation}

Eq.~\ref{eq:ce_logit} shows that cross-entropy can increase target probability either by raising the target logit or by reducing the non-target logit mass through softmax normalization.
Importantly, these two ways of reducing cross-entropy are equivalent from the perspective of target likelihood, but not from the perspective of locality: the latter may alter the relative structure among alternatives that are not intended to be edited.
 Moreover, since one-hot cross-entropy keeps decreasing as the target probability approaches $1$, it continues to reward target amplification even when the target answer is already sufficiently likely. For knowledge editing, this is undesirable: the objective should make the target answer likely enough while preserving output distributions not intended to change. Appendix~\ref{app:ce_gradient} provides the corresponding gradient analysis.

\subsection{Bounded Target Optimization via Logit-odds}
\label{sec:logit_odds}

KLOD replaces unbounded likelihood maximization with bounded target optimization. For $\rho\in(0,1)$, define
\begin{equation}
    \operatorname{logit}(\rho)
    =
    \log\frac{\rho}{1-\rho}.
\end{equation}
At each $t\in\mathcal{T}$, the target logit-odds is
\begin{equation}
    o_t
    =
    \operatorname{logit}
    \left(
    p_\theta(a_t\mid s_{\leq t})
    \right)
    =
    \log
    \frac{
    p_\theta(a_t\mid s_{\leq t})
    }{
    1-p_\theta(a_t\mid s_{\leq t})
    }.
    \label{eq:logit_odds_prob}
\end{equation}
Equivalently,
\begin{equation}
    o_t
    =
    z_{t,a_t}^{\theta}
    -
    \log
    \sum_{v\in\mathcal{V}\setminus\{a_t\}}
    \exp z_{t,v}^{\theta}.
    \label{eq:logit_odds_logit}
\end{equation}
Thus, $o_t$ measures the preference for the target token relative to the aggregate probability mass of non-target tokens. Equivalently, Eq.~\ref{eq:logit_odds_logit} expresses the edit signal as a margin between the target logit and the log-sum-exp of the rest of the vocabulary. This formulation contrasts the target logit with the aggregate non-target logit mass, allowing the desired probability threshold to be expressed directly as a margin constraint.

Let $\alpha\in(0,1]$ denote the configured target probability
threshold. To support the endpoint $\alpha=1$ without numerical
divergence in the logit transform, the implementation uses the
effective threshold
\begin{equation}
    \tilde{\alpha}
    =
    \min(\alpha, 1-\epsilon),
    \qquad
    \epsilon=10^{-6}.
    \label{eq:alpha_clip}
\end{equation}
For all configurations with $\alpha<1$, this clipping has no effect.

Since
\begin{equation}
    p_\theta(a_t\mid s_{\leq t}) \geq \tilde{\alpha}
    \quad\Longleftrightarrow\quad
    o_t \geq \operatorname{logit}(\tilde{\alpha}),
\end{equation}
KLOD uses the one-sided hinge objective
\begin{equation}
    L_{\mathrm{odds}}
    =
    \frac{1}{|\mathcal{T}|}
    \sum_{t\in\mathcal{T}}
    \max
    \left(
        0,
        \operatorname{logit}(\tilde{\alpha})-o_t
    \right).
    \label{eq:odds_loss}
\end{equation}
This loss is active only while the target probability at a prediction position is below the effective threshold, preventing further target amplification after the margin is satisfied. KLOD therefore optimizes for sufficiency rather than maximality:
once the desired target probability is reached, the target objective no longer rewards making the edited answer increasingly dominant.
A thresholded cross-entropy objective can also introduce a stopping mechanism, but it gates the sequence-averaged cross-entropy jointly, whereas KLOD applies a separate margin constraint at each target prediction position. Section~\ref{sec:thresholded_ce} derives their gradient relationship and provides a controlled empirical comparison.

\subsection{Non-target Distribution Preservation}
\label{sec:non_target}

To preserve locality, KLOD allows the intended target probability increase while preserving the relative structure of the non-target distribution. For each $t\in\mathcal{T}$, we exclude the target token $a_t$ and define the renormalized non-target distribution:
\begin{equation}
    q_\theta^t(v)
    =
    \frac{
    \exp z_{t,v}^{\theta}
    }{
    \sum_{u\in\mathcal{V}\setminus\{a_t\}}
    \exp z_{t,u}^{\theta}
    },
    \quad
    v\neq a_t.
    \label{eq:non_target_dist}
\end{equation}
Let $q_{\theta_0}^t$ denote the corresponding distribution of the pre-edit model. KLOD preserves it with
\begin{equation}
    L_{\mathrm{nt}}
    =
    \frac{1}{|\mathcal{T}|}
    \sum_{t\in\mathcal{T}}
    D_{\mathrm{KL}}
    \left(
    q_{\theta_0}^t
    \;\|\;
    q_{\theta}^t
    \right).
    \label{eq:non_target_kl}
\end{equation}

Excluding the target token is essential. A full-distribution KL at target prediction positions would conflict with the intended target update, whereas no constraint would allow arbitrary non-target redistribution. By preserving only the renormalized non-target distribution, $L_{\mathrm{nt}}$ decouples target editing from distribution preservation. Thus, target exclusion avoids directly asking the preservation term to match the probability that the edit objective is designed to change.

\subsection{Prefix Distribution Preservation}
\label{sec:prefix}

Parameter updates can also induce drift at positions that do not directly predict target tokens. KLOD therefore preserves the full next-token distribution at prefix positions:
\begin{equation}
    L_{\mathrm{prefix}}
    =
    \frac{1}{|\mathcal{P}|}
    \sum_{t\in\mathcal{P}}
    D_{\mathrm{KL}}
    \left(
    p_{\theta_0}(\cdot\mid s_{\leq t})
    \;\|\;
    p_{\theta}(\cdot\mid s_{\leq t})
    \right).
    \label{eq:prefix_kl}
\end{equation}
Since prefix positions do not require an intentional target probability increase, KLOD preserves the full vocabulary distribution rather than a non-target distribution.

\subsection{Final Objective and Optimization}
\label{sec:final_objective}

The final KLOD objective is
\begin{equation}
    L_{\mathrm{KLOD}}
    =
    L_{\mathrm{odds}}
    +
    \lambda_{\mathrm{nt}} L_{\mathrm{nt}}
    +
    \lambda_{\mathrm{prefix}} L_{\mathrm{prefix}}
.
    \label{eq:final_objective}
\end{equation}
Here, $\lambda_{\mathrm{nt}}$ and $\lambda_{\mathrm{prefix}}$ control the strength of non-target distribution preservation and prefix distribution preservation.

Following~\citet{LocFT-BF} and \citet{KnowEdit}, we use localized fine-tuning: only the editable parameters in selected layers or modules are updated, while the remaining parameters are frozen. KLOD therefore makes the target answer sufficiently likely up to threshold $\alpha$, while keeping the non-target distribution at target prediction positions and the full next-token distribution at prefix positions close to the pre-edit model.

\section{Experiments}
\label{sec:experiments}

\subsection{Experimental Setup}
\label{sec:exp_setup}

\paragraph{Datasets.}
We evaluate KLOD on ZsRE~\citep{ZsRE} and CounterFact~\citep{ROME} under the Breadth-First optimization protocol of LocFT-BF~\citep{LocFT-BF}. ZsRE formulates factual editing as question answering, whereas CounterFact evaluates counterfactual factual associations. Following LocFT-BF, we sample 3,000 edit requests from each dataset and use the same fixed edit set across methods. Dataset details are provided in Appendix~\ref{app:datasets}.

\paragraph{Models and Editing Setup.}
We use Llama3-8B-Instruct~\citep{Llama3} and Qwen2.5-7B-Instruct~\citep{Qwen2.5}. Under the Breadth-First pipeline, the full edit set is jointly optimized over shuffled mini-batches for multiple epochs. Following LocFT-BF, we update only the MLP down-projection matrix at layer 22 for Llama3-8B-Instruct and layer 6 for Qwen2.5-7B-Instruct. We use $\alpha=0.85$, $\lambda_{\mathrm{nt}}=0.6$, and $\lambda_{\mathrm{prefix}}=1.2$ across all main settings. Hyperparameter selection and implementation details are provided in Appendix~\ref{app:implementation}.

\paragraph{Evaluation Metrics.}
Following standard knowledge editing protocols~\citep{KnowEdit}, we report Reliability, Generalization, and Locality, which measure edit success, rephrase transfer, and preservation of pre-edit predictions on unrelated prompts, respectively. Reliability and Generalization are computed as token-level accuracy under teacher forcing. Recent work has shown that teacher-forced evaluation can overestimate editing performance by exposing gold target information \cite{wild}; we therefore additionally report free-running generation results without teacher forcing in Appendix~\ref{app:free_running}. We additionally report Capability~\citep{ModelEditingHarms,LocFT-BF}, defined as the average performance on GSM8K, MMLU, NQ-Open, SST-2, and WMT16. Formal definitions are provided in Appendix~\ref{app:metrics}.

\paragraph{Baselines.}
We compare KLOD with ROME~\citep{ROME}, AlphaEdit~\citep{AlphaEdit}, UltraEdit~\citep{UltraEdit}, FT-L, FT-M~\citep{KnowEdit}, LocFT-BF~\citep{LocFT-BF}, and OVERTONE~\citep{OVERTONE}. LocFT-BF and OVERTONE provide the most direct fine-tuning-based comparisons. We use a batch size of 100 for methods that support batch editing; baseline-specific settings are provided in Appendix~\ref{app:baselines}.

\begin{table*}[t]
\centering
\resizebox{\textwidth}{!}{%
\begin{tabular}{l|l|c c c c c|c c c c c}
\toprule
\multirow{2}{*}{Model} &
\multirow{2}{*}{Method} &
\multicolumn{5}{c|}{CounterFact} &
\multicolumn{5}{c}{ZsRE} \\
\cmidrule(lr){3-7} \cmidrule(lr){8-12}
 &  & Rel. & Gen. & Loc. & Cap. & Score
 & Rel. & Gen. & Loc. & Cap. & Score \\
\midrule
\multirow{9}{*}{Llama3-8B-Instruct}
& pre-edit  & 0.87  & 1.18  & 100.00 & 62.11 & 41.04 & 26.00 & 25.26 & 100.00 & 62.11 & 53.34 \\
& ROME      & 15.42 & 3.73  & 1.00   & 17.52 & 9.42  & 2.40  & 1.98  & 0.26   & 17.22 & 5.47  \\
& AlphaEdit & 80.33 & \textbf{59.40} & \underline{15.80} & 25.46 & 45.25 & 83.41 & 74.78 & 36.10 & 35.28 & 57.39 \\
& UltraEdit & 35.85 & 18.80 & 6.38   & \textbf{60.80} & 30.46 & 63.45 & 56.23 & 35.21 & 60.63 & 53.88 \\
& FT-L      & 1.37  & 0.40  & 0.00   & 15.07 & 4.21  & 6.96  & 6.58  & 1.03  & 17.88 & 8.11  \\
& FT-M      & 0.67  & 0.65  & 0.65   & 30.99 & 8.24  & 48.84 & 47.61 & 8.06  & 30.10 & 33.65 \\
& LocFT-BF  & \underline{99.77} & 55.83 & 1.87 & 41.92 & 49.85 & \textbf{99.93} & \textbf{89.21} & \underline{51.24} & \textbf{62.10} & \underline{75.62} \\
& OVERTONE  & \textbf{99.80} & \underline{58.12} & 1.83 & 42.51 & \underline{50.57} & \textbf{99.93} & \underline{88.41} & 51.00 & \underline{61.97} & 75.33 \\
& \textbf{KLOD} & 99.70 & 47.37 & \textbf{44.75} & \underline{60.77} & \textbf{63.15} & \underline{99.89} & 86.85 & \textbf{86.47} & 61.42 & \textbf{83.66} \\
\midrule
\multirow{9}{*}{Qwen2.5-7B-Instruct}
& pre-edit  & 1.02  & 0.97  & 100.00 & 54.75 & 39.19 & 26.47 & 25.66 & 100.00 & 54.75 & 51.72 \\
& ROME      & 0.00  & 0.00  & 0.72   & 16.12 & 4.21  & 9.94  & 9.12  & 0.81  & 14.34 & 8.55  \\
& AlphaEdit & 0.00  & 0.00  & 0.07   & 17.53 & 4.40  & 42.82 & 43.73 & 20.38 & 28.76 & 33.92 \\
& UltraEdit & 40.98 & 15.12 & \underline{12.92} & \textbf{54.39} & 30.85 & 58.56 & 49.94 & \underline{74.93} & 55.66 & 59.77 \\
& FT-L      & 19.83 & 11.17 & 1.77   & 14.81 & 11.90 & 3.72  & 3.35  & 1.01  & 16.02 & 6.03  \\
& FT-M      & 0.67  & 0.63  & 0.45   & 32.35 & 8.53  & 37.94 & 34.98 & 6.65  & 18.87 & 24.61 \\
& LocFT-BF  & \underline{99.63} & \underline{55.20} & 8.13 & 50.05 & \underline{53.25} & \underline{99.92} & \textbf{88.56} & 68.72 & \textbf{57.27} & \underline{78.62} \\
& OVERTONE  & \textbf{99.67} & 51.13 & 7.68 & 48.43 & 51.73 & \textbf{100.00} & 87.37 & 68.28 & \underline{56.88} & 78.13 \\
& \textbf{KLOD} & 99.60 & \textbf{68.10} & \textbf{29.45} & \underline{50.48} & \textbf{61.91} & \underline{99.92} & \underline{87.92} & \textbf{78.50} & 56.80 & \textbf{80.79} \\
\bottomrule
\end{tabular}
}
\caption{
Main results on 3k sequential editing. Rel., Gen., Loc., and Cap. denote Reliability, Generalization, Locality, and Capability. Score is their arithmetic mean. Among editing methods, best and second-best results are shown in \textbf{bold} and \underline{underlined}.
}
\label{tab:main_results}
\end{table*}

\subsection{Main Results}
\label{sec:main_results}


Table~\ref{tab:main_results} reports the results of 3k sequential editing on Llama3-8B-Instruct and Qwen2.5-7B-Instruct. Across all model--dataset settings, KLOD achieves the highest Locality among editing methods while maintaining high Reliability.

Several baselines, including ROME, FT-L, FT-M, and AlphaEdit on Qwen2.5-7B-Instruct, show low Reliability, Locality, or Capability, consistent with known instability in sequential or large-scale editing, such as ROME collapse, gradual forgetting, capability degradation, and backbone-dependent behavior~\citep{EditingAtScale,FallOfROME,LocFT-BF}. To clarify that the particularly low FT-M and AlphaEdit results on Qwen2.5-7B-Instruct are associated with degradation under cumulative multi-editing rather than a single anomalous endpoint, we additionally report their performance at 500, 1k, and 3k edits in Appendix~\ref{app:baseline_scale}.

The most direct fine-tuning-based baselines, LocFT-BF and OVERTONE, achieve high Reliability and Generalization but suffer from severe Locality collapse on CounterFact. For Llama3-8B-Instruct, their Locality scores are only 1.87 and 1.83, respectively, whereas KLOD achieves 44.75 Locality while maintaining 99.70 Reliability and 60.77 Capability.
This improvement comes with a real Generalization cost at the primary locality-oriented operating point: KLOD obtains 47.37 Generalization, compared with 55.83 for LocFT-BF and 58.12 for OVERTONE. This should therefore be interpreted as a Generalization--Locality trade-off rather than a uniform improvement over all metrics. The behavior is setting-dependent: on Qwen2.5-7B-Instruct CounterFact, KLOD improves both Generalization and Locality over the two direct fine-tuning baselines, reaching 68.10 and 29.45, respectively.
Despite the large relative improvement, absolute CounterFact Locality remains limited after 3k edits: 44.75 for Llama3-8B-Instruct and 29.45 for Qwen2.5-7B-Instruct. KLOD therefore mitigates, rather than eliminates, accumulated locality degradation in this setting.

On ZsRE, KLOD also substantially improves Locality while maintaining high Reliability, with relatively small Generalization gaps to the strongest fine-tuning baselines. For Llama3-8B-Instruct, KLOD obtains 99.89 Reliability and 86.47 Locality, with Generalization remaining close to the strongest fine-tuning baselines.
For Qwen2.5-7B-Instruct, KLOD achieves the best overall Score, with 99.92 Reliability, 87.92 Generalization, and 78.50 Locality.
These results show that KLOD provides a more favorable Rel.--Gen.--Loc. balance than standard fine-tuning-based editing objectives, especially when fine-tuning baselines obtain high edit success at the cost of collapsed Locality.
Together, these results show that KLOD provides controllable Reliability--Generalization--Locality operating points rather than uniformly dominating every metric at a single setting.

\begin{table*}[t]
\centering
\resizebox{\textwidth}{!}{%
\begin{tabular}{l|ccc|ccc|cc}
\toprule
\multirow{2}{*}{Method} &
\multicolumn{3}{c|}{Rewrite} &
\multicolumn{3}{c|}{Rephrase} &
\multicolumn{2}{c}{Locality} \\
\cmidrule(lr){2-4} \cmidrule(lr){5-7} \cmidrule(lr){8-9}
& Prompt KL & Target KL & Non-target KL
& Prompt KL & Target KL & Non-target KL
& Prompt KL & Target KL \\
\midrule
ROME      & 16.597 & 14.312 & 14.308 & 15.377 & 13.522 & 13.511 & 15.143 & 13.911 \\
AlphaEdit & 3.782  & 8.418  & 5.596  & 3.583  & 6.537  & 5.061  & 3.535  & 4.809  \\
UltraEdit & 3.470  & 4.962  & 4.650  & 1.180  & 3.578  & 3.459  & 3.218  & 4.521  \\
LocFT-BF  & 2.697  & 15.936 & 8.967  & 2.296  & 10.118 & 8.124  & 2.481  & 10.135 \\
OVERTONE  & 2.635  & 15.266 & 8.680  & 2.181  & 9.698  & 7.794  & 2.378  & 9.729  \\
\textbf{KLOD}
          & \textbf{0.031} & \textbf{4.942} & \textbf{0.147}
          & \textbf{0.150} & \textbf{1.763} & \textbf{0.579}
          & \textbf{0.135} & \textbf{1.126} \\
\bottomrule
\end{tabular}
}
\caption{
Distributional KL divergence after 3k sequential editing on CounterFact with Llama3-8B-Instruct. Lower values indicate smaller drift from the pre-edit model.
}
\label{tab:distribution_kl}
\end{table*}

\subsection{Distribution KL Analysis}
\label{sec:distribution_kl}

To examine whether the preservation terms in KLOD indeed reduce output distribution drift, we measure the KL divergence between the pre-edit model and the edited model after 3k sequential editing on CounterFact with Llama3-8B-Instruct. Table~\ref{tab:distribution_kl} reports distributional drift in rewrite, rephrase, and locality contexts. The rewrite context is the prompt to which the edit is directly applied, the rephrase context evaluates the same edit under a paraphrased prompt, and the locality context is unrelated to the edit but requires preserving pre-edit behavior. In each context, Prompt KL measures drift in the next-token distribution at prompt-side positions before predicting the target answer, while Target KL measures full-vocabulary distribution drift at target answer positions. For rewrite and rephrase contexts, we also report Non-target KL, computed after excluding the target token and renormalizing the remaining distribution. Detailed position definitions, calculation examples in Appendix~\ref{app:kl_analysis}, and additional KL analysis on Qwen2.5-7B-Instruct are provided in Appendix~\ref{app:qwen_kl_analysis}.

KLOD achieves the lowest KL at all measured positions, indicating that the proposed preservation objective effectively reduces distributional drift. In particular, KLOD obtains Non-target KL scores of 0.147 and 0.579 in the rewrite and rephrase contexts, respectively, substantially lower than OVERTONE and LocFT-BF. This shows that the non-target distribution is preserved even as the target token probability increases. KLOD also achieves the lowest Prompt KL in rewrite, rephrase, and locality contexts, with scores of 0.031, 0.150, and 0.135, respectively, demonstrating that the prefix preservation term suppresses prompt-level changes in next-token behavior. Overall, this KL analysis supports the interpretation that KLOD's locality gains arise not from weakening the edit, but from suppressing unnecessary distributional drift in target and prompt contexts.

An instance-level analysis in Appendix~\ref{app:kl_distribution} further shows that these reductions persist from the median through the 99th percentile, rather than being restricted to the token-averaged mean.

\subsection{Optimization Stability Across Seeds}
\label{sec:seed_stability}

\begin{table}[t]
\centering
\small
\resizebox{\columnwidth}{!}{%
\begin{tabular}{llccc}
\toprule
Dataset & Method & Rel. & Gen. & Loc. \\
\midrule
\multirow{3}{*}{CounterFact}
& KLOD      & $99.72 \pm 0.02$ & $46.76 \pm 1.01$ & $\mathbf{46.51 \pm 1.98}$ \\
& LocFT-BF  & $99.74 \pm 0.02$ & $56.83 \pm 0.95$ & $1.89 \pm 0.18$ \\
& OVERTONE  & $99.73 \pm 0.07$ & $57.56 \pm 0.90$ & $1.77 \pm 0.05$ \\
\midrule
\multirow{3}{*}{ZsRE}
& KLOD      & $99.90 \pm 0.02$ & $87.39 \pm 0.47$ & $\mathbf{86.38 \pm 0.33}$ \\
& LocFT-BF  & $99.81 \pm 0.11$ & $89.57 \pm 0.32$ & $51.15 \pm 0.20$ \\
& OVERTONE  & $99.78 \pm 0.16$ & $88.75 \pm 0.30$ & $51.25 \pm 0.25$ \\
\bottomrule
\end{tabular}
}
\caption{
Mean $\pm$ standard deviation over three Breadth-First training seeds
with Llama3-8B-Instruct. The sampled 3k edit set and all
hyperparameters are fixed.
}
\label{tab:seed_stability}
\end{table}

Table~\ref{tab:main_results} reports the seed-42 run used for the full method comparison and subsequent analyses.
To evaluate optimization stability, we additionally repeat KLOD and its two closest fine-tuning-based baselines, LocFT-BF and OVERTONE, using seeds $\{1,42,100\}$ on both datasets with Llama3-8B-Instruct.
The sampled 3k edit set, pretrained model, editable layer, and all hyperparameters are held fixed; only the random seed, which controls stochastic factors such as epoch-wise mini-batch shuffling, is varied.

Unlike online edit-by-edit optimization, the Breadth-First protocol jointly optimizes the full edit set over multiple shuffled epochs. The experiment therefore evaluates the stability of the resulting stochastic optimization trajectory rather than permutations of a fixed online request sequence.

 Table~\ref{tab:seed_stability} summarizes the three-seed results. KLOD's Locality advantage remains stable across seeds.
On CounterFact, KLOD obtains $46.51\pm1.98$ Locality, whereas LocFT-BF and OVERTONE remain near 2.
On ZsRE, KLOD obtains $86.38\pm0.33$ Locality, compared with approximately 51 for both baselines.
The lower CounterFact Generalization of KLOD also persists across seeds, confirming that the observed Generalization--Locality trade-off is systematic rather than run-to-run noise.

\subsection{Effect of Target Probability Threshold}
\label{sec:alpha_sweep}

\begin{figure}[t]
    \centering
    \includegraphics[width=\linewidth]{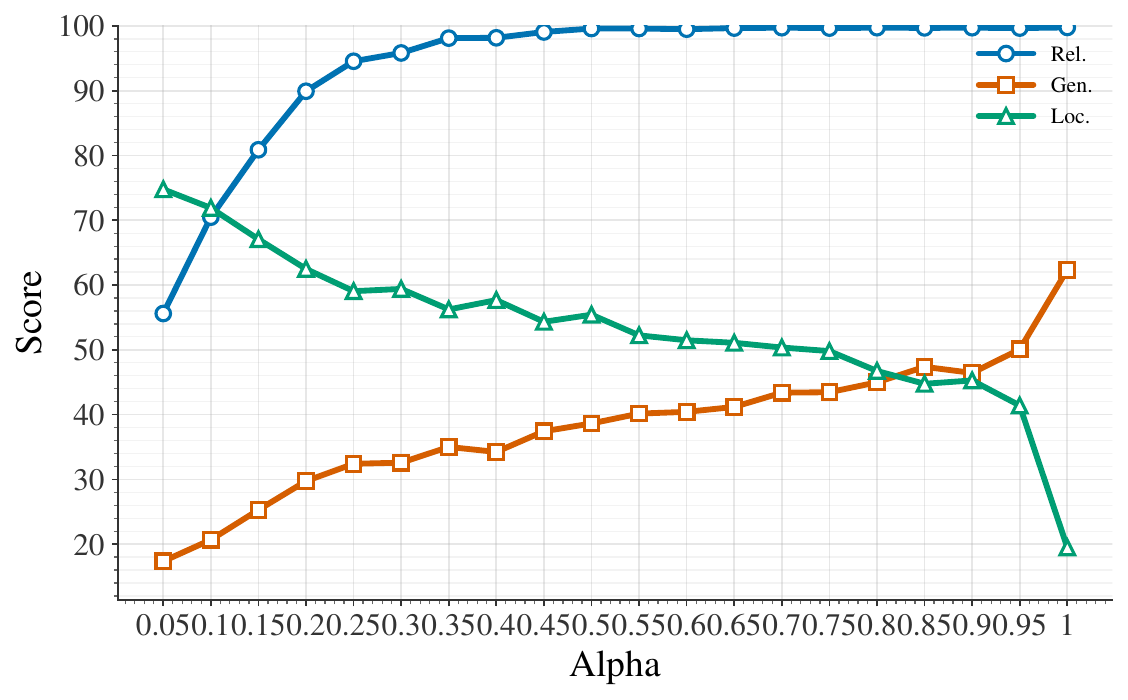}
        \caption{
        Generalization--Locality trade-off across target probability
        thresholds $\alpha$ on CounterFact with Llama3-8B-Instruct.
        The $\alpha=1.0$ endpoint is clipped to $1-10^{-6}$ when computing
        the logit-odds margin.
        }
    \label{fig:alpha_sweep}
\end{figure}

Figure~\ref{fig:alpha_sweep} shows the effect of the target probability threshold $\alpha$ on KLOD's edit behavior. In the target logit-odds margin objective, $\alpha$ determines the desired probability level that the target token should reach. A smaller $\alpha$ imposes a weaker constraint on the target edit, whereas a larger $\alpha$ encourages a stronger edit by requiring a higher target probability. Empirically, Reliability is low when $\alpha$ is small, indicating that edits do not succeed reliably when the target probability threshold is insufficiently high ($\alpha < 0.4$). In contrast, when $\alpha > 0.5$, most edits succeed, as the target token is trained to receive more than half of the probability mass.

As $\alpha$ increases, Generalization improves while Locality decreases, exposing an explicit Generalization--Locality operating curve. The primary configuration $\alpha=0.85$ is locality-oriented, achieving 47.37 Generalization and 44.75 Locality on Llama3/CounterFact. At the configured high-Generalization endpoint $\alpha=1.0$, Generalization increases to 62.35, while Locality decreases to 19.48 and Capability remains 57.65. This operating point exceeds the Generalization of both LocFT-BF (55.83) and OVERTONE (58.12), while retaining substantially higher Locality and Capability than either baseline.
We therefore use $\alpha=0.85$ as the primary operating point because it prioritizes locality preservation while maintaining near-perfect Reliability, rather than because it uniformly maximizes all metrics. Sensitivity analyses for the preservation weights are provided in Appendix~\ref{app:lambda_sweep}.

\subsection{Ablation Study}
\label{sec:ablation}

\begin{table}[t]
\centering
\small
\begin{tabular}{lcccc}
\toprule
Method & Rel. & Gen. & Loc. & Cap. \\
\midrule
KLOD & 99.70 & 47.37 & 44.75 & 60.77 \\
w/o $L_{\mathrm{prefix}}$ (w/ $L_{\mathrm{nt}}$) & 99.60 & 52.05 & 41.30 & 60.70 \\
w/o $L_{\mathrm{prefix}}$ (w/ $L_{\mathrm{full}}$) & 99.73 & 53.72 & 32.90 & 57.72 \\
w/o $L_{\mathrm{nt}}$ & 99.70 & 52.70 & 9.15 & 60.78 \\
w/o $L_{\mathrm{prefix}},L_{\mathrm{nt}}$ & 99.72 & 61.73 & 2.02 & 44.24 \\
w/o All & 99.77 & 55.83 & 1.87 & 41.92 \\
\bottomrule
\end{tabular}
\caption{
Ablation study on CounterFact with Llama3-8B-Instruct.
$L_{\mathrm{full}}$ denotes a full-vocabulary KL at target positions, including the target token, used in place of $L_{\mathrm{nt}}$.
w/o All corresponds to LocFT-BF with standard cross-entropy.
}
\label{tab:ablation_counterfact}
\end{table}

Table~\ref{tab:ablation_counterfact} presents the ablation results for KLOD's target objective and preservation terms. All variants maintain Reliability above 99\%, indicating that each variant still induces the target edit sufficiently. Comparing the variant without both preservation terms to LocFT-BF, the bounded logit-odds objective slightly improves Locality from 1.87 to 2.02 and Capability from 41.92 to 44.24. However, these gains are limited, suggesting that KLOD's locality improvement mainly comes from the KL-based preservation terms. A controlled no-prefix comparison that independently varies the
target objective and target-position KL distribution is reported in Appendix~\ref{app:controlled_ablation} (Table~\ref{tab:controlled_ablation}).

In particular, $L_{\mathrm{nt}}$ is not merely a KL regularizer, but a key design for decoupling target editing from distribution preservation. When $L_{\mathrm{prefix}}$ is removed, KLOD with $L_{\mathrm{nt}}$ still maintains 41.30 Locality and 60.70 Capability. In contrast, replacing $L_{\mathrm{nt}}$ with $L_{\mathrm{full}}$, a full-vocabulary KL that also includes the target token, reduces Locality to 32.90 and Capability to 57.72. This shows that including the target token in the KL term at target positions prevents the intended target update from being cleanly separated from the preservation objective. By contrast, KLOD's $L_{\mathrm{nt}}$ preserves only the renormalized non-target distribution, explicitly decoupling the target edit induced by $L_{\mathrm{odds}}$ from non-target distribution preservation.

The relative contributions of the two preservation terms are dataset-dependent. On CounterFact, $L_{\mathrm{nt}}$ is the dominant preservation component: removing it reduces Locality from 44.75 to 9.15, whereas removing $L_{\mathrm{prefix}}$ reduces Locality only to 41.30. Thus, target-position non-target preservation is particularly important in the CounterFact setting. The pattern is more balanced on ZsRE. As shown in Appendix~\ref{sec:ablation_zsre}, using $L_{\mathrm{nt}}$ without $L_{\mathrm{prefix}}$ yields 81.26 Locality, while using $L_{\mathrm{prefix}}$ without $L_{\mathrm{nt}}$ yields 82.09; combining both reaches 86.47. Thus, the two preservation terms have complementary roles on ZsRE. A complementary post-hoc decomposition of locality prediction changes in Appendix~\ref{app:locality_decomposition} (Table~\ref{tab:locality_decomposition}) further characterizes
how the failure pattern differs between CounterFact and ZsRE.

\begin{figure}[t]
    \centering
    \includegraphics[width=0.90\columnwidth]
    {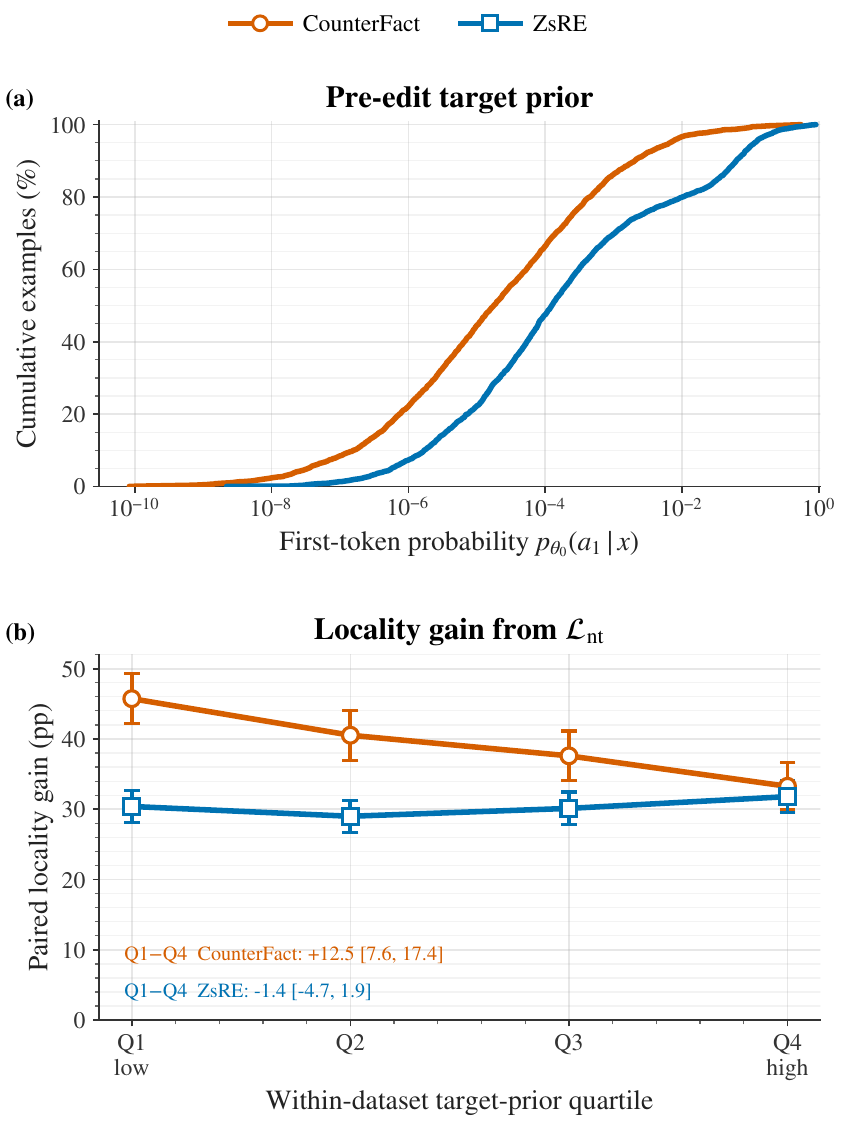}
    \caption{
    Effect of pre-edit target prior on $L_{\mathrm{nt}}$.
    (a) Distribution of $p_{\theta_0}(a_1\mid x)$.
    (b) Mean paired Locality gain across target-prior quartiles.
    Error bars and Q1--Q4 intervals are normal-approximation 95\% CIs
    ($n=750$ per quartile).
    }
    \label{fig:target_prior_effect}
\end{figure}

To further investigate this dataset-dependent behavior, we analyze the pre-edit probability assigned to the first edited target token, $p_{\theta_0}(a_1\mid x)$, where $x$ is the rewrite prompt. As shown in Figure~\ref{fig:target_prior_effect}, CounterFact contains substantially more low-prior targets than ZsRE.

We group each dataset into quartiles according to this pre-edit target prior and measure the case-level paired Locality gain from adding $L_{\mathrm{nt}}$ to the bounded target objective. On CounterFact, the gain decreases from $+45.7$ percentage points in the lowest-prior quartile (Q1) to $+33.3$ points in the highest-prior quartile (Q4), yielding a Q1--Q4 difference of $+12.5$ points (normal-approximation 95\% CI: $[7.6,17.4]$). In contrast, ZsRE shows no clear quartile-dependent difference ($-1.4$ points; normal-approximation 95\% CI: $[-4.7,1.9]$). The intervals reflect case-level uncertainty, with 750 cases in each quartile.

This suggests that target-side non-target preservation is particularly beneficial when the intended target initially receives weak support from the pre-edit model. We treat this as supporting evidence for the dataset-dependent role of $L_{\mathrm{nt}}$, rather than as a causal explanation of all differences between CounterFact and ZsRE.

\subsection{Comparison with Thresholded Cross-Entropy}
\label{sec:thresholded_ce}

A natural alternative to the logit-odds hinge is to introduce a
stopping mechanism directly into cross-entropy. We consider
sequence-level thresholded cross-entropy,
\begin{equation}
    L_{\mathrm{TCE}}
    =
    \max\left(
        0,
        L_{\mathrm{CE}} + \log \alpha
    \right),
    \label{eq:tce_main}
\end{equation}
where $L_{\mathrm{CE}}$ is the mean target-position negative
log-likelihood defined in Eq.~\ref{eq:ce}.

Although both objectives bound target optimization, they differ in how the edit signal is applied. Thresholded CE gates the sequence-averaged objective jointly, whereas KLOD applies a separate logit-odds margin constraint at each target prediction position. Moreover, while active, the cross-entropy gradient at a target position is scaled by $1-p_t$ relative to the corresponding logit-odds gradient. Thus, the CE signal becomes progressively weaker as the target probability increases, whereas the logit-odds hinge maintains its margin signal until the position-specific threshold is satisfied. Appendix~\ref{app:thresholded_ce} provides the detailed derivation.

For the matched-Generalization comparison, we keep $\alpha$, the editable layer, optimizer, batch size, number of epochs, and all other optimization settings fixed. We vary only the preservation coefficients of the thresholded-CE variant and report the operating point whose Generalization is closest to that of the primary KLOD configuration.

At a matched-Generalization operating point, KLOD achieves 44.75 Locality compared with 37.50 for thresholded CE, while maintaining comparable Reliability and Generalization (Table~\ref{tab:thresholded_ce_main}). This suggests that KLOD's advantage is not explained solely by introducing a stopping threshold: the position-wise logit-odds formulation provides a more locality-preserving optimization signal.

\begin{table}[t]
\centering
\small
\begin{tabular}{lccc}
\toprule
Target objective & Rel. & Gen. & Loc. \\
\midrule
Thresholded CE
& 99.80 & 46.20 & 37.50 \\
Logit-odds hinge (KLOD)
& 99.70 & 47.37 & \textbf{44.75} \\
\bottomrule
\end{tabular}
\caption{
Matched-Generalization comparison between KLOD and sequence-level
thresholded cross-entropy on CounterFact with Llama3-8B-Instruct.
}
\label{tab:thresholded_ce_main}
\end{table}

\subsection{Scalability to Large-Scale Sequential Editing}
\label{sec:scale_analysis}

\begin{figure*}[t]
    \centering
    \includegraphics[width=\textwidth]{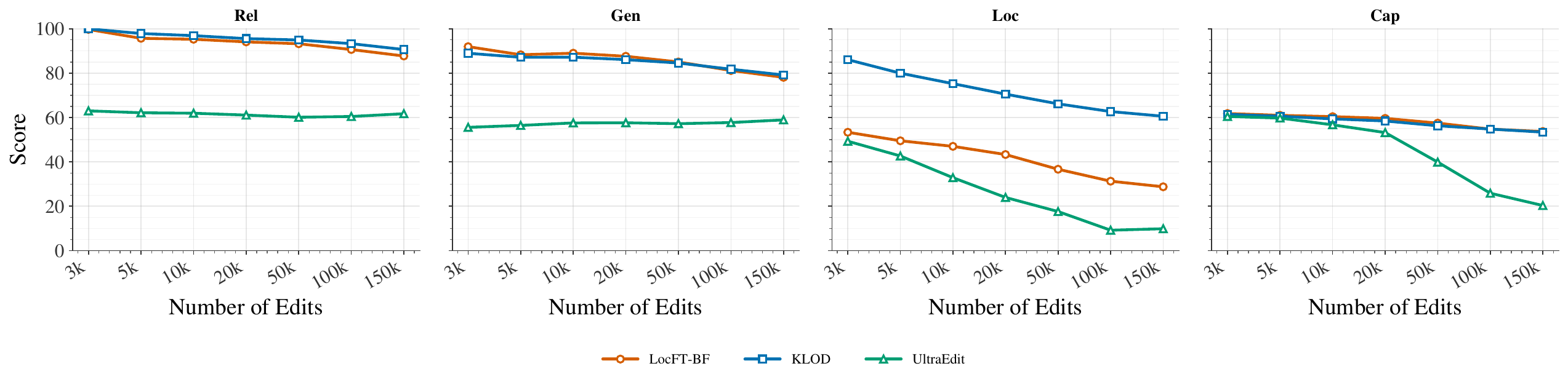}
    \caption{
    Scaling behavior on ZsRE from 3k to 150k sequential edits. KLOD maintains Rel. and Gen. comparable to LocFT-BF while substantially improving Loc. across edit scales with Llama3-8B-Instruct.
    }
    \label{fig:zsre_150k}
\end{figure*}

Figure~\ref{fig:zsre_150k} shows sequential editing performance on ZsRE as the number of edits increases from 3k to 150k. These scaling results provide key evidence that KLOD's locality gain is not simply due to edit weakening. Across edit scales, KLOD maintains Reliability and Generalization comparable to LocFT-BF. Reliability remains high at all scales, reaching around 90 even after 150k edits. Generalization also follows a similar gradual decline to LocFT-BF as the number of edits increases, while still maintaining competitive rephrase generalization at 150k. This indicates that KLOD's distribution preservation terms mitigate locality degradation from accumulated edits without interfering with the target edit itself.

The largest difference appears in Locality. As the number of edits increases, LocFT-BF and UltraEdit show steady Locality degradation, dropping to roughly 30 and 10, respectively, at 150k edits. In contrast, KLOD achieves over 85 Locality at 3k edits and maintains around 60 even at 150k edits. For Capability, KLOD remains comparable to LocFT-BF, whereas UltraEdit shows substantial capability degradation as the edit scale grows. These results show that KLOD effectively mitigates distributional drift and locality degradation not only in small-scale editing but also in large-scale sequential editing, while preserving high Reliability and competitive Generalization.

\begin{table}[t]
\centering
\small
\begin{tabular}{lcccc}
\toprule
\multirow{2}{*}{Method} &
\multicolumn{2}{c}{Llama3-8B} &
\multicolumn{2}{c}{Qwen2.5-7B} \\
\cmidrule(lr){2-3} \cmidrule(lr){4-5}
& CounterFact & ZsRE & CounterFact & ZsRE \\
\midrule
ROME  & 7.68 & 8.46 & 6.71 & 8.50 \\
FT-L  & 1.60 & 1.70 & 0.34 & 1.00 \\
LocFT-BF & 0.16 & 0.23 & 0.17 & 0.30 \\
KLOD  & 0.37 & 0.55 & 0.43 & 0.62 \\
\bottomrule
\end{tabular}
\caption{
Average editing time per edit in the 3k sequential editing setting. KLOD adds modest overhead over LocFT-BF while remaining under one second per edit.
}
\label{tab:editing_time}
\end{table}

\subsection{Editing Time}
\label{sec:editing_time}

Table~\ref{tab:editing_time} reports the average editing time per edit request in seconds under the 3k sequential editing setting. KLOD requires 0.37--0.55 seconds on Llama3-8B-Instruct and 0.43--0.62 seconds on Qwen2.5-7B-Instruct, editing at a much lower cost than locate-and-edit methods such as ROME. Although KLOD is slower than LocFT-BF, the additional overhead mainly comes from computing the reference-model distributions required by the non-target and prefix KL terms. Nevertheless, KLOD remains below one second per edit across all model--dataset settings, showing that it maintains practical sequential editing cost while adding distribution preservation.

\section{Conclusion}
\label{sec:conclusion}

This work studies an objective-level mismatch in fine-tuning-based knowledge editing: standard cross-entropy increases the edited target probability without explicitly constraining changes in the non-target output distribution. We propose KLOD, which combines bounded target optimization with target-excluded non-target preservation and prefix distribution preservation to better separate the intended edit from changes that should remain limited.

Across CounterFact and ZsRE, KLOD substantially mitigates locality degradation while maintaining high Reliability, although the results also reveal a clear Generalization--Locality trade-off in some settings. The target probability threshold provides an explicit way to control this operating point. Ablation and distributional KL analyses further support the interpretation that KLOD's locality gains are associated with preserving output distributions rather than simply reducing edit strength.

More broadly, our results suggest that fine-tuning-based knowledge editing can benefit from treating an edit as a selective distributional update: the objective should encourage the intended target change while explicitly constraining distributions that are not meant to change. KLOD provides one practical realization of this principle for sequential editing.

\section*{Limitations}
\label{sec:limitations}

KLOD does not eliminate the inherent trade-off among edit strength, Generalization, and Locality, and its operating point may vary across datasets and backbones. Despite substantial relative improvements, absolute CounterFact Locality remains limited after 3k edits (44.75 on Llama3-8B-Instruct and 29.45 on Qwen2.5-7B-Instruct), showing that KLOD mitigates rather than eliminates locality degradation.
Our primary evaluation follows established CounterFact and ZsRE protocols, supplemented by a 3k-edit WikiBigEdit experiment in Appendix~\ref{app:wikibigedit}. These protocols' token-level locality metrics may not fully capture broader behavioral preservation. Although the additional free-running evaluation in Appendix~\ref{app:free_running} shows similar behavior without teacher forcing, broader generation-based evaluation across datasets and editing settings remains necessary. KLOD also incurs additional computation over standard localized fine-tuning due to its reference-distribution KL terms.



\section*{Acknowledgments}
This work was supported by a National Research Foundation of Korea (NRF) grant funded by the Government of Korea (MSIT) (No. 2022R1A2C1005316).

\bibliography{custom}
\appendix




\section{Related Work}
\label{app:related_work}

\subsection{Locate-and-Edit Paradigm}

A major line of knowledge editing follows the locate-and-edit paradigm, which identifies and directly modifies internal computations or parameter regions responsible for factual associations. ROME uses causal tracing to identify feed-forward modules important for factual recall and modifies factual associations through rank-one updates~\cite{ROME}. AlphaEdit focuses on the observation that parameter perturbations in sequential editing can damage preserved knowledge, and proposes a projection-based editing strategy~\cite{AlphaEdit}. These methods are primarily designed around where the target knowledge is localized, such as specific layers, modules, or subspaces, and how the corresponding parameter update should be applied.

Our work studies fine-tuning-based knowledge editing from a complementary perspective. We do not aim to identify new parameter regions responsible for factual associations or design a closed-form update rule. Instead, given a set of editable parameters, we analyze what training objective should be optimized when these parameters are updated through gradient-based optimization. In particular, we focus on an objective-level issue: increasing the likelihood of the target answer for edit success may unnecessarily redistribute the output distribution outside the target token. Thus, unlike locate-and-edit methods that mainly focus on edit location, KLOD focuses on distribution-preserving objective design that enables localized fine-tuning to achieve target editing and locality preservation simultaneously.

\subsection{Fine-Tuning-Based Knowledge Editing}

Fine-tuning-based editing is architecture-agnostic and simple to implement, but it has long been regarded as a disruptive baseline in knowledge editing because gradient updates can affect broader model behavior beyond the target fact. Recent studies, however, show that the reliability and locality of fine-tuning-based editing are not determined solely by the update mechanism itself, but depend heavily on the objective, editable parameter selection, and optimization protocol. \citet{StandardFT} show that standard fine-tuning can achieve competitive editing performance through conditional likelihood and locality training. LocFT-BF improves the stability of fine-tuning-based editing in large-scale sequential editing through localized tuning and Breadth-First mini-batch optimization~\cite{LocFT-BF}.

These studies suggest that fine-tuning should not be viewed merely as a weak baseline, but as an optimization framework that can be redesigned for knowledge editing. However, existing fine-tuning-based editing methods mainly focus on how reliably the target answer can be learned or how edit requests should be optimized. In contrast, we focus on the fact that standard cross-entropy-based fine-tuning does not explicitly preserve the distributional structure among non-target tokens while increasing target probability. Cross-entropy is effective at increasing target token probability, but it does not distinguish whether this increase comes from raising the target logit or from changing the non-target logit mass and the relative distribution among non-target tokens. Since knowledge editing requires not only making the target answer sufficiently likely but also keeping behavior unrelated to the target close to the pre-edit model, such non-target redistribution can accumulate as distributional drift and locality degradation in sequential editing.

Recent work such as OVERTONE and Editing Overfit shows that naive target maximization can induce objective-level side effects, including token-level overfitting and excessive target probability assignment~\cite{OVERTONE,EditingOverfit}. These studies are closely related to ours in that they analyze how the fine-tuning objective affects edit behavior. However, KLOD goes beyond mitigating target-side overfitting or stabilizing target likelihood optimization by directly controlling non-target distribution drift during target probability increase. In other words, KLOD is not concerned only with how strongly the model should predict the target token, but with how the non-target distribution should be preserved while the target token is edited.

KLOD is also distinct from simple KL-regularized fine-tuning. Applying KL regularization to the full output distribution at target positions can directly conflict with the edit objective, since the edited target token must increase in probability while the reference distribution encourages the original distribution to be preserved. To avoid this conflict, KLOD excludes the target token at target positions, defines a renormalized non-target distribution over the vocabulary excluding the target token, and keeps this distribution close to that of the pre-edit model. This target-excluded preservation allows the intended target probability increase while suppressing unnecessary redistribution in the non-target distribution. In addition, KLOD uses a bounded logit-odds objective that stops further target amplification once the target probability reaches a specified threshold, thereby mitigating over-editing caused by unbounded likelihood maximization. Thus, while KLOD follows the broader line of fine-tuning-based KE, it differs from prior methods by explicitly decoupling target editing from non-target distribution preservation through a distribution-preserving objective.

\subsection{Sequential Editing and Side Effects}

In practical knowledge editing, edits accumulate over time, making stability under sequential or lifelong editing as important as the success of a single edit. Continual editing and Editing-at-Scale studies show that as the number of edits increases, models may forget previous edits or suffer degradation in downstream task performance~\cite{ContinualEditing,EditingAtScale}. It has also been reported that model editing can harm general abilities such as reasoning, natural language inference, and question answering~\cite{ModelEditingHarms}. These findings indicate that knowledge editing should be evaluated not only by Reliability, but also by side effects such as Generalization, Locality, Retention, and Capability.

This work addresses such side effects at the output-distribution level. While prior sequential editing studies mainly observe and mitigate task-level forgetting, locality degradation, and capability drop caused by edit accumulation, KLOD approaches these issues from the perspective that such degradation can be linked to distributional drift permitted by the fine-tuning objective. In particular, standard cross-entropy is not designed to preserve the distributional structure outside the target token while increasing the target answer probability. As many edits accumulate, small non-target redistribution may therefore accumulate and appear as locality degradation. KLOD aims to mitigate this accumulated output distribution drift while maintaining edit success by keeping the non-target distribution at target positions and the next-token distribution at prefix positions close to those of the pre-edit model. In this sense, KLOD does not treat the side effects of sequential editing only as task-level metrics, but directly intervenes in distribution-level changes induced by the fine-tuning objective.

\section{Gradient of Softmax Cross-Entropy}
\label{app:ce_gradient}

This section summarizes the learning signal of softmax cross-entropy discussed in Section~\ref{sec:ce_revisit} from a gradient perspective.
Consider a single prediction position $t$, with gold next token $a_t$ and logits $z_t^\theta$.
The cross-entropy loss is
\begin{equation}
    \ell_t
    =
    -\log p_\theta(a_t\mid s_{\leq t})
    =
    -z_{t,a_t}^{\theta}
    +
    \log\sum_{v\in\mathcal{V}}\exp z_{t,v}^{\theta}.
\end{equation}
Given the softmax probability
\begin{equation}
    p_\theta(v\mid s_{\leq t})
    =
    \frac{\exp z_{t,v}^{\theta}}
    {\sum_{u\in\mathcal{V}}\exp z_{t,u}^{\theta}},
\end{equation}
the gradient with respect to each vocabulary logit is
\begin{equation}
    \frac{\partial \ell_t}
    {\partial z_{t,v}^{\theta}}
    =
    p_\theta(v\mid s_{\leq t})
    -
    \mathbb{I}[v=a_t].
    \label{eq:ce_gradient}
\end{equation}
Thus, for the target token $a_t$,
\begin{equation}
    \frac{\partial \ell_t}
    {\partial z_{t,a_t}^{\theta}}
    =
    p_\theta(a_t\mid s_{\leq t})-1,
\end{equation}
whereas for a non-target token $v\neq a_t$,
\begin{equation}
    \frac{\partial \ell_t}
    {\partial z_{t,v}^{\theta}}
    =
    p_\theta(v\mid s_{\leq t}).
\end{equation}

In logit space, gradient descent therefore increases the target logit while assigning non-target logits a decreasing signal proportional to their current model probabilities.
This gradient structure is effective for increasing the target token probability, but it does not require the relative structure of the non-target distribution to remain close to that of the pre-edit model.
In other words, cross-entropy directly optimizes target probability increase, but does not explicitly control how the probability mass outside the target token is redistributed.
Under shared parameter updates and sequential editing, such unconstrained non-target redistribution can accumulate as distributional drift and locality degradation.

To mitigate this issue, KLOD's non-target preservation term keeps the renormalized non-target distribution, excluding the target token, close to the pre-edit model.
This allows the intended increase in target probability while directly suppressing unnecessary changes in the non-target distribution.

\section{Relation to Thresholded Cross-Entropy}
\label{app:thresholded_ce}

This section compares the position-wise logit-odds hinge used by KLOD
with sequence-level thresholded cross-entropy introduced in
Section~\ref{sec:thresholded_ce}.

\paragraph{Stopping condition.}
For a target position $t\in\mathcal{T}$, let
\begin{equation}
    p_t
    =
    p_\theta(a_t\mid s_{\leq t}).
\end{equation}
Recall that the target-sequence cross-entropy is
\begin{equation}
    L_{\mathrm{CE}}
    =
    -\frac{1}{|\mathcal{T}|}
    \sum_{t\in\mathcal{T}}
    \log p_t,
\end{equation}
and thresholded cross-entropy is defined as
\begin{equation}
    L_{\mathrm{TCE}}
    =
    \max
    \left(
        0,
        L_{\mathrm{CE}}+\log\alpha
    \right).
\end{equation}

Thresholded CE becomes inactive when
\begin{equation}
    L_{\mathrm{CE}}
    \leq
    -\log\alpha.
\end{equation}
Equivalently,
\begin{equation}
    \left(
        \prod_{t\in\mathcal{T}} p_t
    \right)^{1/|\mathcal{T}|}
    \geq
    \alpha.
\end{equation}
Thus, thresholded CE applies the stopping condition to the geometric
mean of the target probabilities over the sequence.

In KLOD, the target logit-odds at position $t$ is
\begin{equation}
    o_t
    =
    \log\frac{p_t}{1-p_t}.
\end{equation}
The position-wise hinge becomes inactive when
\begin{equation}
    o_t \geq \operatorname{logit}(\tilde{\alpha}),
\end{equation}
which is equivalent to
\begin{equation}
    p_t \geq \tilde{\alpha}.
\end{equation}
Therefore, each target position stops contributing to
$L_{\mathrm{odds}}$ once its own threshold is satisfied, while other
positions can remain active.

For $\alpha<1$, the two stopping conditions coincide for a single-token target. For the configured endpoint $\alpha=1$, KLOD uses the clipped threshold $\tilde{\alpha}=1-\epsilon$.
For multi-token targets, however, thresholded CE uses a sequence-level
condition, whereas KLOD applies the threshold independently at each
target position.

\paragraph{Gradient relationship.}
We next compare the learning signal below the threshold.
Consider a target position $t$ for which the KLOD hinge is active.
The position-wise logit-odds loss is
\begin{equation}
    \ell_{\mathrm{odds},t}
    =
    \operatorname{logit}(\tilde{\alpha})-o_t.
\end{equation}
Using the renormalized non-target distribution
\begin{equation}
    q_t(v)
    =
    \frac{
        \exp z_{t,v}^{\theta}
    }{
        \sum_{u\neq a_t}
        \exp z_{t,u}^{\theta}
    },
    \qquad
    v\neq a_t,
\end{equation}
its gradients are
\begin{equation}
    \frac{\partial \ell_{\mathrm{odds},t}}
    {\partial z_{t,a_t}^{\theta}}
    =
    -1,
\end{equation}
and
\begin{equation}
    \frac{\partial \ell_{\mathrm{odds},t}}
    {\partial z_{t,v}^{\theta}}
    =
    q_t(v),
    \qquad
    v\neq a_t.
\end{equation}

For cross-entropy,
\begin{equation}
    \ell_{\mathrm{CE},t}
    =
    -\log p_t,
\end{equation}
and the target-logit gradient is
\begin{equation}
    \frac{\partial \ell_{\mathrm{CE},t}}
    {\partial z_{t,a_t}^{\theta}}
    =
    p_t-1
    =
    -(1-p_t).
\end{equation}
For a non-target token $v\neq a_t$,
\begin{equation}
    \frac{\partial \ell_{\mathrm{CE},t}}
    {\partial z_{t,v}^{\theta}}
    =
    p_\theta(v\mid s_{\leq t})
    =
    (1-p_t)q_t(v).
\end{equation}

Therefore, at an active target position, the cross-entropy gradient has
the same direction as the logit-odds gradient, but
its magnitude is scaled by $1-p_t$.
As the target probability increases, the cross-entropy signal becomes
progressively weaker, whereas the logit-odds hinge maintains its
margin signal until the position-specific threshold is reached.

In summary, both objectives can bound target optimization, but they
differ in two respects: thresholded CE uses sequence-level stopping
with probability-dependent gradient scaling, whereas KLOD uses
position-wise stopping with an unattenuated margin signal at active
target positions.
The corresponding matched-Generalization comparison is reported in
Table~\ref{tab:thresholded_ce_main}.

\section{Dataset Details}
\label{app:datasets}

\paragraph{ZsRE.}
Each edit request in ZsRE consists of a rewrite question and a target answer, along with rephrased questions and locality prompts. We use the rewrite question as the edit input and the rephrased questions for Generalization evaluation. The locality prompts are used to measure whether the pre-edit model's prediction is preserved on questions unrelated to the edit. Thus, ZsRE evaluates both whether edited knowledge generalizes to semantically equivalent expressions and whether behavior on unrelated questions is preserved.

\paragraph{CounterFact.}
Each edit request in CounterFact consists of a rewrite prompt and a target object, along with paraphrase prompts and locality prompts. The rewrite prompt is used for Reliability evaluation, the paraphrase prompts for Generalization evaluation, and the locality prompts for evaluating Locality, which measures pre-edit behavior preservation. Thus, CounterFact evaluates both whether the target fact is incorporated and whether unrelated facts are preserved.

\section{Implementation Details}
\label{app:implementation}

KLOD is implemented under the localized fine-tuning with Breadth-First pipeline setting of LocFT-BF~\cite{LocFT-BF}. The Breadth-First pipeline is not an online-style sequential optimization procedure that fully optimizes each edit request before moving to the next. Instead, it jointly optimizes the full edit set through repeated mini-batch updates over multiple epochs. This optimization protocol is used to mitigate overwriting among sequential edits.

In localized fine-tuning, we do not update all parameters of the backbone model. Instead, only the MLP down-projection matrix in a specified editable layer is made trainable. Specifically, we edit the MLP down-projection matrix at layer 22 for Llama3-8B-Instruct and at layer 6 for Qwen2.5-7B-Instruct. Attention modules, embedding layers, layer normalization parameters, and all other MLP parameters are kept frozen.

Each edit example is constructed by concatenating the edit prompt \(x\) and target response \(y=(y_1,\ldots,y_m)\) as \(s=x\oplus y\). In the standard target-only fine-tuning objective, negative log-likelihood is computed only at positions \(\mathcal{T}\) that predict the target response, while prompt positions and padding positions are masked out from NLL computation. Thus, the prompt \(x\) is used only as the conditioning context for target response generation, and no direct NLL supervision is applied to the prompt tokens themselves.

KLOD keeps the same LocFT-BF optimization environment but extends the standard target-only cross-entropy objective into a distribution-preserving objective. Specifically, at target prediction positions \(\mathcal{T}\), KLOD computes the bounded logit-odds loss and the target-excluded non-target KL. At valid prefix positions \(\mathcal{P}\), excluding target prediction positions, KLOD adds the prefix KL to preserve the next-token distribution of the pre-edit model. Padding positions are excluded from all objective terms.

The objective hyperparameters of KLOD are fixed across all main experiments. We set the target-token probability threshold to \(\alpha=0.85\), and the loss weights for non-target distribution preservation and prefix distribution preservation to \(\lambda_{\mathrm{nt}}=0.6\) and \(\lambda_{\mathrm{prefix}}=1.2\), respectively. We use the Adam optimizer. For each mini-batch, gradients are computed only for the selected MLP down-projection matrix, followed by a single optimizer update. Backbone-specific hyperparameters, including learning rate, batch size, number of epochs, and weight decay, are summarized in Table~\ref{tab:implementation_hparams}.
 All experiments were conducted using three NVIDIA GeForce RTX 3090 GPUs. This hardware environment was used for both the main editing experiments and the editing-time measurements.

\subsection{Hyperparameter Selection}
\label{app:hparam_selection}

\begin{table}[t]
\centering
\small
\resizebox{\columnwidth}{!}{%
\begin{tabular}{ccc|ccc}
\toprule
$\alpha$
& $\lambda_{\mathrm{prefix}}$
& $\lambda_{\mathrm{nt}}$
& Rel. & Gen. & Loc. \\
\midrule
0.85 & 0.8 & 0.6 & 100.00 & 87.66 & 88.35 \\
0.85 & 1.2 & 0.4 & 100.00 & 88.04 & 88.51 \\
0.75 & 1.2 & 0.6 & 100.00 & 85.03 & 88.60 \\
\textbf{0.85} & \textbf{1.2} & \textbf{0.6}
& \textbf{100.00} & \textbf{87.77} & \textbf{88.48} \\
0.95 & 1.2 & 0.6 & 100.00 & 89.75 & 88.56 \\
0.85 & 1.2 & 0.8 & 99.95 & 85.50 & 88.86 \\
0.85 & 1.6 & 0.6 & 100.00 & 86.82 & 88.35 \\
\bottomrule
\end{tabular}
}
\caption{
One-at-a-time sensitivity analysis on the held-out 500-example ZsRE
set.
The bold row denotes the shared configuration used in the main
experiments.
}
\label{tab:hparam_validation}
\end{table}

We select the shared KLOD configuration using a held-out
500-example ZsRE set that is disjoint from the 3k requests used in
the main experiments.
We first exclude the main 3k requests from the larger ZsRE pool and
then sample 500 non-overlapping requests using random seed 42.
The held-out requests are edited and evaluated using their rewrite,
rephrase, and locality fields following the same knowledge-editing
protocol as the main experiments.

Table~\ref{tab:hparam_validation} reports a small one-at-a-time sensitivity analysis over
$\alpha$, $\lambda_{\mathrm{nt}}$, and
$\lambda_{\mathrm{prefix}}$.
Rather than maximizing a single metric, we choose
$\alpha=0.85$, $\lambda_{\mathrm{nt}}=0.6$, and
$\lambda_{\mathrm{prefix}}=1.2$ as a balanced, non-extreme
operating point and fix this configuration across all four primary
model--dataset settings.

\begin{table}[t]
\centering
\small
\begin{tabular}{lcc}
\toprule
Hyperparameter & Llama3-8B & Qwen2.5-7B \\
\midrule
Editable layer & 22 & 6 \\
Editable module & MLP down-proj & MLP down-proj \\
$\alpha$ & 0.85 & 0.85 \\
$\lambda_{\mathrm{nt}}$ & 0.6 & 0.6 \\
$\lambda_{\mathrm{prefix}}$ & 1.2 & 1.2 \\
Batch size & 100 & 100 \\
Epochs & 25 & 25 \\
Learning rate & 5.0e-4 & 5.0e-4 \\
Weight decay & 0 & 0 \\
\bottomrule
\end{tabular}
\caption{Implementation hyperparameters for KLOD.}
\label{tab:implementation_hparams}
\end{table}

\section{Evaluation Metrics}
\label{app:metrics}

We evaluate the edited model after sequential editing using four metrics: Reliability, Generalization, Locality, and Capability. Following the traditional teacher-forcing evaluation protocol in EasyEdit~\cite{easyedit}, evaluation is based on token-level prediction accuracy at target positions rather than answer-level exact match. For each edit request $i$, let $x_i$ denote the rewrite prompt, $x_i^{\mathrm{reph}}$ the rephrased prompt, and $y_i=(y_{i,1},\ldots,y_{i,m_i})$ the target answer token sequence. We denote the final edited model by $f_{\theta_N}$ and the pre-edit model by $f_{\theta_0}$.

Given a prompt $x$ and a target token sequence $y=(y_1,\ldots,y_m)$, we define the prediction of model $f_\theta$ at the $t$-th target position as
\begin{equation}
    \hat{y}^{\theta}_{t}(x,y)
    =
    \arg\max_{v\in\mathcal{V}}
    p_{\theta}(v \mid x, y_{<t}).
\end{equation}
The corresponding token-level accuracy is
\begin{equation}
    \mathrm{Acc}_{\theta}(x,y)
    =
    \frac{1}{m}
    \sum_{t=1}^{m}
    \mathbb{I}
    \left[
    \hat{y}^{\theta}_{t}(x,y)=y_t
    \right].
\end{equation}
Thus, for multi-token answers, the score is computed by the fraction of correctly predicted tokens, rather than requiring the entire string to match exactly.

\paragraph{Reliability.}
Reliability measures how accurately the edited model predicts the target answer for the rewrite prompt:
\begin{equation}
    \mathrm{Rel.}
    =
    \frac{1}{|\mathcal{I}_{\mathrm{rw}}|}
    \sum_{i\in\mathcal{I}_{\mathrm{rw}}}
    \mathrm{Acc}_{\theta_N}(x_i^{\mathrm{rw}},y_i).
\end{equation}

\paragraph{Generalization.}
Generalization measures whether the edit transfers to rephrased prompts:
\begin{equation}
    \mathrm{Gen.}
    =
    \frac{1}{|\mathcal{I}_{\mathrm{reph}}|}
    \sum_{i\in\mathcal{I}_{\mathrm{reph}}}
    \mathrm{Acc}_{\theta_N}(x_i^{\mathrm{reph}},y_i).
\end{equation}

\paragraph{Locality.}
Locality evaluates whether the pre-edit model's prediction is preserved on locality prompts unrelated to the edit. Let $x_i^{\mathrm{loc}}$ be a locality prompt and $y_i^{\mathrm{loc}}=(y_{i,1}^{\mathrm{loc}},\ldots,y_{i,\ell_i}^{\mathrm{loc}})$ be the reference continuation. The locality prediction of model $f_\theta$ is defined as
\begin{equation}
    \hat{y}^{\theta,\mathrm{loc}}_{i,t}
    =
    \arg\max_{v\in\mathcal{V}}
    p_{\theta}(v \mid x_i^{\mathrm{loc}}, y_{i,<t}^{\mathrm{loc}}).
\end{equation}
Locality is computed as the fraction of token-level predictions that match between the edited model and the pre-edit model:
\begin{equation}
    \mathrm{Loc.}
    =
    \frac{1}{|\mathcal{I}_{\mathrm{loc}}|}
    \sum_{i\in\mathcal{I}_{\mathrm{loc}}}
    \frac{1}{\ell_i}
    \sum_{t=1}^{\ell_i}
    \mathbb{I}
    \left[
    \hat{y}^{\theta_N,\mathrm{loc}}_{i,t}
    =
    \hat{y}^{\theta_0,\mathrm{loc}}_{i,t}
    \right].
\end{equation}
Thus, Locality measures how well the original behavior is preserved on locality prompts, rather than whether the edited model predicts a specific target answer.

\paragraph{Capability.}
Capability evaluates the model's general abilities after sequential editing by averaging performance on GSM8K, MMLU, NQ-Open, SST-2, and WMT16:
\begin{equation}
    \mathrm{Cap.}
    =
    \frac{1}{|\mathcal{B}|}
    \sum_{b\in\mathcal{B}}
    \mathrm{Score}_b(f_{\theta_N}),
\end{equation}
where $\mathcal{B}$ denotes the benchmark set consisting of GSM8K, MMLU, NQ-Open, SST-2, and WMT16. The task-specific score for each benchmark is converted to a $[0,100]$ scale before averaging.

\paragraph{Overall Score.}
The Score reported in the main results is the arithmetic mean of the four metrics:
\begin{equation}
    \mathrm{Score}
    =
    \frac{
    \mathrm{Rel.}
    +
    \mathrm{Gen.}
    +
    \mathrm{Loc.}
    +
    \mathrm{Cap.}
    }{4}.
\end{equation}

Because these metrics capture competing editing desiderata, the unweighted average is reported only as a descriptive summary and is not used to select the operating point of KLOD.

\subsection{Reliability on Multi-token Targets}
\label{app:multitoken_reliability}

Because $L_{\mathrm{nt}}$ excludes the current gold next token at each
target prediction position, each token of a multi-token target is
excluded separately when it becomes the gold token at its own
autoregressive step.
We additionally evaluate Reliability exclusively on examples whose
targets contain more than one tokenizer token. 

In addition to token-level Reliability, we report a stricter
example-level criterion that requires every target position to predict
the gold token as top-1 under teacher forcing:
\begin{equation}
    \mathrm{StrictRel}_i
    =
    \mathbb{I}
    \left[
    \forall t \in \mathcal{T}_i,\;
    \hat{y}_{i,t}=y_{i,t}
    \right].
\end{equation}


\begin{table}[t]
\centering
\small
\resizebox{\columnwidth}{!}{%
\begin{tabular}{llccc}
\toprule
Model & Dataset & Multi-token targets
& Token Rel. & Strict Rel. \\
\midrule
Llama3-8B & ZsRE
& 2,313/3,000 (77.1\%) & 99.85 & 99.61 \\
Qwen2.5-7B & ZsRE
& 2,313/3,000 (77.1\%) & 99.90 & 99.57 \\
Llama3-8B & CounterFact
& 40/3,000 (1.3\%) & 100.00 & 100.00 \\
Qwen2.5-7B & CounterFact
& 40/3,000 (1.3\%) & 100.00 & 100.00 \\
\bottomrule
\end{tabular}
}
\caption{
Reliability evaluated exclusively on multi-token target examples.
Strict Rel. requires all target positions in an example to predict the
gold token correctly.
}
\label{tab:multitoken_reliability}
\end{table}

As shown in Table~\ref{tab:multitoken_reliability}, even on ZsRE, where 77.1\% of the targets are multi-token, strict
all-token Reliability remains above 99.5\%.
This indicates that the position-wise target exclusion used by
$L_{\mathrm{nt}}$ does not prevent reliable completion of multi-token
targets under the reported evaluation protocol.

\section{Baseline Details}
\label{app:baselines}

This section describes the implementation details and hyperparameter settings of the baseline editors used in our experiments.
ROME, AlphaEdit, UltraEdit, FT-L, and FT-M are evaluated based on the implementations released in EasyEdit.
For these methods, we use the public configurations provided by EasyEdit and do not perform additional hyperparameter tuning on the test set.
Model forward computation and evaluation are performed in bfloat16 precision whenever possible.
However, for methods such as ROME and AlphaEdit that rely on method-specific linear algebra operations, including covariance statistics, matrix inversion, and projection, we compute the corresponding statistics and linear algebra components in float32 precision or follow the implementation default precision for numerical stability.
LocFT-BF is evaluated using the official implementation released by the authors, with hyperparameters reported in the original paper.
Since OVERTONE does not provide an official public implementation, we reimplement it based on the method described in the paper.
ROME, FT-L, and FT-M use a batch size of 1, while the remaining baselines use a batch size of 100.
All methods are evaluated using the same backbone model, tokenizer, sampled edit set, and evaluation protocol. Each method follows its designated optimization protocol, and model parameters are not reset during cumulative editing. For methods that operate edit-by-edit, updates are applied cumulatively without model reinitialization, whereas LocFT-BF, OVERTONE, and KLOD use the Breadth-First optimization protocol described in Section~\ref{sec:exp_setup}.

\paragraph{ROME.}
ROME is a locate-and-edit method that identifies internal computations associated with factual associations through causal tracing and modifies a single factual association by applying a rank-one update to selected feed-forward weights~\citep{ROME}.
We use the ROME implementation and hyperparameter configuration released in EasyEdit.
For both Llama3-8B-Instruct and Qwen2.5-7B-Instruct, we edit the MLP down-projection module at layer 5.
For both models, the subject token is set to \texttt{subject\_last}, and value vector optimization is performed for 25 epochs.
The learning rate is set to $5\times10^{-1}$, weight decay to $10^{-3}$, clamp norm factor to 4, and KL factor to 0.0625.
Covariance statistics are computed from 100,000 Wikipedia samples, with the covariance dtype kept in float32.
Value vector optimization follows the default precision of the EasyEdit implementation, while model forward computation uses bfloat16 precision whenever possible.

\paragraph{AlphaEdit.}
AlphaEdit addresses the issue that parameter perturbations introduced during locate-then-edit procedures can damage preserved knowledge.
To mitigate this, it projects edit perturbations onto the null space of preserved knowledge before applying the update, aiming to reduce locality degradation in sequential editing~\citep{AlphaEdit}.
We use the AlphaEdit implementation and hyperparameter configuration released in EasyEdit.
For both Llama3-8B-Instruct and Qwen2.5-7B-Instruct, we edit the MLP down-projection modules in layers 4--8, with layer selection set to \texttt{all} and the fact token set to \texttt{subject\_last}.
The batch size is set to 100.
The null-space projection threshold is set to $2\times10^{-2}$, and the $L_2$ coefficient is set to 1.
For Llama3-8B-Instruct, the value optimization learning rate is set to $1\times10^{-1}$, weight decay to 0.5, and clamp norm factor to 0.75.
For Qwen2.5-7B-Instruct, the value optimization learning rate is set to $5\times10^{-1}$, weight decay to $10^{-3}$, and clamp norm factor to 4.
For both models, value optimization is performed for 25 epochs, and the KL factor is set to 0.0625.
Covariance-related statistics are computed from 100,000 Wikipedia samples in float32 precision.
Method-specific linear algebra operations, including null-space projection, use float32 or the implementation default precision for numerical stability, while model forward computation is performed in bfloat16 precision whenever possible.

\paragraph{FT-L and FT-M.}
FT-L and FT-M are fine-tuning-based editing baselines~\citep{KnowEdit}.
FT-L fine-tunes localized parameters using supervision on the target answer token at the last token position of the prompt.
FT-M applies cross-entropy loss to the full target answer while masking prompt tokens, making the objective closer to standard target-answer fine-tuning.
We construct FT-L and FT-M based on the FT implementation released in EasyEdit.
In the EasyEdit configuration, we use the setting where \texttt{objective\_optimization} is \texttt{prompt\_last} as FT-L, and the setting where it is \texttt{target\_new} as FT-M.
For Llama3-8B-Instruct and Qwen2.5-7B-Instruct, we fine-tune the MLP down-projection weights at layer 21 and layer 27, respectively.
Optimization is performed for 25 epochs, with a learning rate of $5\times10^{-4}$ and a maximum sequence length of 40 in the hyperparameter configuration.
The batch size is set to 1 for both FT-L and FT-M.
Weight decay and KL factor are both set to 0, and no norm constraint is used.
Model forward and backward computation are performed in bfloat16 precision by default, while optimizer states and other training components follow the implementation default precision.

\paragraph{UltraEdit.}
UltraEdit is a training-free, subject-free, and memory-free editing method designed for scalable lifelong editing.
It computes parameter shifts in one step using lightweight linear algebra operations and employs lifelong normalization, which updates feature statistics as editing turns accumulate, to improve stability in large-scale lifelong editing~\citep{UltraEdit}.
We use the UltraEdit implementation and hyperparameter configuration released in EasyEdit.
For Llama3-8B-Instruct, we edit the MLP gate-projection in layers 11--15 and the MLP up-projection in layers 18--24.
For Qwen2.5-7B-Instruct, we edit the MLP gate-projection and up-projection in layers 18--26.
For both models, the learning rate is set to $1\times10^{-6}$, and the token setting is set to \texttt{mask}.
The batch size, one-pass batch size, and editor batch size are set to 100, 10, and 1024, respectively.
Model forward computation is performed in bfloat16 precision whenever possible, while method-specific linear algebra operations follow the default precision of the EasyEdit implementation.

\paragraph{LocFT-BF.}
LocFT-BF argues that the weak performance of prior fine-tuning-based editors may stem not from an inherent limitation of fine-tuning itself, but from implementation choices such as depth-first sample-wise optimization and suboptimal tuning locations.
Based on this analysis, LocFT-BF combines Breadth-First mini-batch optimization with localized tuning location selection to provide a fine-tuning-based editing baseline better suited for sequential editing~\citep{LocFT-BF}.
We use the official implementation released by the authors.
For Llama3-8B-Instruct and Qwen2.5-7B-Instruct, we tune the MLP down-projection modules at layer 22 and layer 6, respectively.
Optimization is performed for 25 epochs, with a learning rate of $5\times10^{-4}$, batch size of 100, and weight decay of 0.
Model forward and backward computation are performed in bfloat16 precision, while optimizer states and other training components follow the default precision of the official implementation.

\paragraph{OVERTONE.}
OVERTONE identifies heterogeneous token overfitting (HTO), where tokens in the target sequence overfit at different rates during knowledge editing.
To mitigate this issue, it uses a token-level smoothing objective that adaptively adjusts the target distribution according to each token's fitting state~\citep{OVERTONE}.
Since OVERTONE does not provide an official public implementation, we reimplement it based on the objective and reported hyperparameters in the original paper. 

The implementation is built on the same localized fine-tuning pipeline as LocFT-BF and KLOD.
For an objective-level comparison, we keep the editable layer, editable module, batch size, learning rate, number of epochs, sampled edit set, Breadth-First optimization protocol, decoding, and evaluation setting identical.
We do not separately tune OVERTONE for the test performance of each dataset or backbone.
The batch size is set to 100, and the OVERTONE-related coefficients are fixed to $\lambda=0.1$, $\epsilon=0.01$, and $n_{\sigma}=0.5$, following the original paper.
Model forward and backward computation are performed in bfloat16 precision by default, while optimizer states and other training components follow the implementation default precision.

\begin{table}[t]
\centering
\small
\resizebox{\columnwidth}{!}{%
\begin{tabular}{lrrrrrr}
\toprule
Method & Edits & Rel. & Gen. & Loc. & Cap. & Score \\
\midrule
\multirow{3}{*}{FT-M}
& 500 & 100.00 & 99.90 & 1.50 & 22.70 & 56.02 \\
& 1k  & 99.95  & 99.75 & 1.30 & 15.33 & 54.08 \\
& 3k  & 0.67   & 0.63  & 0.45 & 32.35 & 8.53 \\
\midrule
\multirow{3}{*}{AlphaEdit}
& 500 & 82.10 & 43.70 & 20.20 & 19.71 & 41.43 \\
& 1k  & 88.05 & 55.30 & 20.10 & 18.87 & 45.58 \\
& 3k  & 0.00  & 0.00  & 0.07  & 17.53 & 4.40 \\
\midrule
\multirow{3}{*}{KLOD}
& 500 & 99.80 & 56.30 & 44.90 & 52.51 & 63.38 \\
& 1k  & 99.70 & 63.45 & 34.10 & 52.80 & 62.51 \\
& 3k  & 99.60 & 68.10 & 29.45 & 50.48 & 61.91 \\
\bottomrule
\end{tabular}%
}
\caption{
Cumulative edit-scale analysis on CounterFact with
Qwen2.5-7B-Instruct.
}
\label{tab:baseline_scale}
\end{table}

\section{Cumulative Edit-Scale Analysis of FT-M and AlphaEdit}
\label{app:baseline_scale}

Table~\ref{tab:baseline_scale} reports cumulative edit-scale results for FT-M, AlphaEdit, and KLOD. To clarify the low 3k results of FT-M and AlphaEdit, we evaluate their behavior as the number of cumulative edits increases on Qwen2.5-7B-Instruct and CounterFact. The pre-edit Capability score is 54.75.

FT-M retains near-perfect edit success at 500 and 1k edits, but its
Locality is already close to zero and Capability is substantially below
the pre-edit value.
At 3k edits, its Reliability and Generalization also collapse.
AlphaEdit likewise exhibits substantial degradation on this backbone
under cumulative editing and nearly collapses by 3k edits.
This behavior is backbone-dependent: with the same implementation,
AlphaEdit remains substantially more reliable on Llama3-8B-Instruct
in Table~\ref{tab:main_results}.
In contrast, KLOD maintains near-perfect Reliability and substantially
higher Locality and Capability across the three edit scales.

The non-monotonic Capability values of FT-M should not be interpreted
as recovery, since Capability averages heterogeneous downstream tasks
and the 3k score remains far below the pre-edit model while Reliability,
Generalization, and Locality have collapsed.

\begin{table*}[t]
\centering
\resizebox{\textwidth}{!}{%
\begin{tabular}{l|ccc|ccc|cc}
\toprule
\multirow{2}{*}{Method} &
\multicolumn{3}{c|}{Rewrite} &
\multicolumn{3}{c|}{Rephrase} &
\multicolumn{2}{c}{Locality} \\
\cmidrule(lr){2-4} \cmidrule(lr){5-7} \cmidrule(lr){8-9}
& Prompt KL & Target KL & Non-target KL
& Prompt KL & Target KL & Non-target KL
& Prompt KL & Target KL \\
\midrule
ROME      & 8.019  & 7.917  & 7.440  & 8.651  & 8.071  & 7.521  & 8.095  & 6.430  \\
AlphaEdit & 12.919 & 12.981 & 12.949 & 12.669 & 12.157 & 12.130 & 13.293 & 12.962 \\
UltraEdit & 0.702  & 6.030  & 1.690  & \textbf{0.228} & 3.999 & \textbf{1.218} & \textbf{0.188} & 2.985 \\
LocFT-BF  & 4.106  & 16.573 & 10.200 & 2.672  & 9.980  & 8.491  & 3.463  & 10.746 \\
OVERTONE  & 3.959  & 16.391 & 10.259 & 2.482  & 9.646  & 8.381  & 3.396  & 10.863 \\
\textbf{KLOD}
          & \textbf{0.111} & \textbf{4.035} & \textbf{0.552}
          & 0.462 & \textbf{2.712} & 1.508
          & 0.261 & \textbf{1.538} \\
\bottomrule
\end{tabular}
}
\caption{
Distributional KL divergence after 3k sequential editing on CounterFact with Qwen2.5-7B-Instruct.
Lower values indicate smaller distributional drift from the pre-edit model.
Best values are shown in \textbf{bold}.
We use the same evaluation protocol as Table~\ref{tab:distribution_kl}.
}
\label{tab:distribution_kl_qwen}
\end{table*}

\section{Details of Distribution KL Analysis}
\label{app:kl_analysis}

This section describes the implementation details of the distribution KL analysis used in Section~\ref{sec:distribution_kl}.
Each evaluation item consists of a prompt $x$ and a target sequence $y$, and we compare the next-token distributions of the pre-edit model $f_{\theta_0}$ and the edited model $f_{\theta}$ on the teacher-forced sequence $s=x\oplus y$.
Rewrite and rephrase items use the edit prompt or rephrased prompt together with the edited target answer, respectively, while locality items use a locality prompt unrelated to the edit and its corresponding ground-truth answer.

KL is computed at shifted prediction positions in causal language modeling.
That is, the logits at position $t$ are treated as the distribution for predicting the next token $s_{t+1}$.
Prompt KL is computed at prompt-side positions that predict tokens within the prompt, while the position that predicts the first target token after observing the final prompt token is included in Target KL.
Target KL measures full-vocabulary distribution drift at positions that predict target answer tokens under teacher forcing.
All KL values are computed in the pre-edit-to-edited direction:
\begin{equation}
    D_{\mathrm{KL}}
    \left(
    p_{\theta_0}(\cdot \mid s_{\leq t})
    \,\|\,
    p_{\theta}(\cdot \mid s_{\leq t})
    \right).
\end{equation}

\paragraph{Position split example.}
For example, suppose the teacher-forced sequence is
\begin{equation}
    s = [x_1, x_2, x_3, y_1, y_2].
\end{equation}
Under shifted prediction in causal language modeling, $x_1$ predicts $x_2$, $x_1,x_2$ predict $x_3$, and $x_1,x_2,x_3$ predict $y_1$.
Therefore, the positions predicting $x_2$ and $x_3$ are included in Prompt KL, whereas the position predicting the first target token $y_1$ is included in Target KL.
In other words, even if the final prompt token is used as context, the corresponding position is treated as a target prediction position if it predicts the first target token.

For rewrite and rephrase contexts, we also compute Non-target KL at target positions.
Let $a_t$ denote the gold next token at each target position.
We exclude this token from the vocabulary and renormalize the distribution over the remaining tokens:
\begin{equation}
    q_{\theta}^{t}(v)
    =
    \frac{
    p_{\theta}(v \mid s_{\leq t})
    }{
    1 - p_{\theta}(a_t \mid s_{\leq t})
    },
    \quad
    v \in \mathcal{V}\setminus\{a_t\}.
\end{equation}
Non-target KL is computed as the KL divergence between $q_{\theta_0}^{t}$ and $q_{\theta}^{t}$, measuring relative distribution drift among non-target tokens rather than changes in the target probability itself.
For locality contexts, we report only Prompt KL and Target KL, since the goal is not to analyze the non-target distribution around the edited target answer.

The final reported values are token-level averages rather than example-level averages.
For each context type, we accumulate the position-wise KL sum and the number of evaluated tokens, and compute the values in Table~\ref{tab:distribution_kl} by dividing the total KL sum by the total token count.

\begin{table*}[t]
\centering
\small
\begin{tabular}{llrrrrr}
\toprule
Method & Component & Mean & Median & P90 & P95 & P99 \\
\midrule
\multirow{2}{*}{KLOD}
& Prompt
& \textbf{0.099} & \textbf{0.063} & \textbf{0.226}
& \textbf{0.306} & \textbf{0.513} \\
& Non-target
& \textbf{0.364} & \textbf{0.168} & \textbf{0.808}
& \textbf{1.342} & \textbf{3.119} \\
\midrule
\multirow{2}{*}{LocFT-BF}
& Prompt
& 2.475 & 2.324 & 3.957 & 4.535 & 5.588 \\
& Non-target
& 8.567 & 8.369 & 12.671 & 14.234 & 16.718 \\
\midrule
\multirow{2}{*}{OVERTONE}
& Prompt
& 2.379 & 2.225 & 3.776 & 4.312 & 5.366 \\
& Non-target
& 8.257 & 8.015 & 12.350 & 13.930 & 16.298 \\
\midrule
\multirow{2}{*}{UltraEdit}
& Prompt
& 2.686 & 2.386 & 5.078 & 5.829 & 7.603 \\
& Non-target
& 4.065 & 3.709 & 6.860 & 7.921 & 9.992 \\
\bottomrule
\end{tabular}
\caption{
Instance-level KL distributions after 3k sequential edits on
CounterFact with Llama3-8B-Instruct.
Prompt aggregates rewrite, rephrase, and locality contexts, whereas
Non-target aggregates target-position KL from rewrite and rephrase
contexts after target-token exclusion and renormalization.
Lower values indicate less distributional drift.
}
\label{tab:kl_distribution}
\end{table*}

\section{Additional Distribution KL Results}
\label{app:qwen_kl_analysis}

Table~\ref{tab:distribution_kl_qwen} reports additional distribution KL analysis after 3k sequential editing on CounterFact with Qwen2.5-7B-Instruct.
The overall trend is similar to the results on Llama3-8B-Instruct.
KLOD achieves the lowest Prompt KL, Target KL, and Non-target KL in the rewrite context, and it also yields the lowest Target KL in the rephrase and locality contexts.
In particular, LocFT-BF and OVERTONE show high Target KL and Non-target KL in the rewrite and rephrase contexts, whereas KLOD substantially reduces both.
This indicates that KLOD's KL-based preservation terms effectively suppress target-side distributional drift even while increasing the target probability.

UltraEdit achieves the lowest Prompt KL in the rephrase and locality contexts, preserving some prompt-side next-token behavior well.
It also obtains the lowest Rephrase Non-target KL.
However, KLOD achieves lower Target KL in both the rephrase and locality contexts, as well as lower Rewrite Non-target KL.
This suggests that KLOD more consistently preserves the target-position distribution in contexts directly tied to the edited answer, whereas UltraEdit is more effective at reducing prompt-side drift in some non-rewrite contexts.
Therefore, KLOD shows consistent advantages in target-side distribution preservation also on Qwen2.5-7B-Instruct.

\section{Instance-level KL Distribution Analysis}
\label{app:kl_distribution}

The main KL analysis reports token-averaged mean divergence.
Although this summarizes overall distributional drift, a mean alone
may be influenced by the shape of the KL distribution.
We therefore additionally examine instance-level KL distributions to
determine whether KLOD's reduction in distributional drift persists
beyond the average.

For each evaluation instance, we first average the position-wise KL
values within that instance.
We then compute the mean, median, 90th percentile (P90), 95th
percentile (P95), and 99th percentile (P99) over instances.
For Prompt KL, we aggregate instances from the rewrite, rephrase, and
locality contexts.
For Non-target KL, we aggregate the target positions from the rewrite
and rephrase contexts, where the target token is excluded and the
remaining distribution is renormalized as in
Section~\ref{app:kl_analysis}.

Table~\ref{tab:kl_distribution} shows that KLOD's reduction in
distributional drift is consistent across the reported KL
distribution, rather than appearing only in the mean.
For Prompt KL, KLOD obtains a median of 0.063 and a P99 of 0.513,
whereas the other methods have median values above 2.2 and P99 values
above 5.3.
The same pattern is stronger for Non-target KL:
KLOD obtains a median of 0.168 and a P99 of 3.119, compared with
medians of 8.369 and 8.015 and P99 values of 16.718 and 16.298 for
LocFT-BF and OVERTONE, respectively.

Notably, KLOD's P99 Non-target KL remains lower than even the median
Non-target KL of LocFT-BF and OVERTONE.
Thus, the substantially lower KL observed for KLOD is not explained
by the token-averaged mean alone; the reduction persists from the
center of the instance-level distribution to its high-KL tail.
This provides additional evidence that KLOD broadly suppresses
output-distribution drift rather than obtaining a low average from a
small subset of low-drift cases.

\section{Controlled Target-Objective and Preservation Ablation}
\label{app:controlled_ablation}

To isolate the effects of bounded target optimization and target
exclusion in the preservation term, we conduct a controlled no-prefix
comparison on CounterFact with Llama3-8B-Instruct.
All variants omit $L_{\mathrm{prefix}}$ and independently vary the
target objective and the target-position KL distribution.

\begin{table}[t]
\centering
\small
\resizebox{\columnwidth}{!}{%
\begin{tabular}{llcccc}
\toprule
Target objective & Preservation & Rel. & Gen. & Loc. & Cap. \\
\midrule
CE & Full KL
& 99.77 & 54.22 & 22.55 & 51.47 \\
CE & Non-target KL
& 99.77 & 54.42 & 26.77 & 54.11 \\
Logit-odds hinge & Full KL
& 99.73 & 53.72 & 32.90 & 57.72 \\
Logit-odds hinge & Non-target KL
& 99.60 & 52.05 & \textbf{41.30} & \textbf{60.70} \\
\bottomrule
\end{tabular}
}
\caption{
Controlled no-prefix ablation on CounterFact with
Llama3-8B-Instruct.
}
\label{tab:controlled_ablation}
\end{table}

Table~\ref{tab:controlled_ablation} separates the contribution of
target exclusion from that of the bounded target objective.
With cross-entropy, replacing full-vocabulary KL with target-excluded
non-target KL improves Locality from 22.55 to 26.77.
The same replacement under the logit-odds hinge improves Locality from
32.90 to 41.30.
Likewise, replacing cross-entropy with the bounded logit-odds objective
improves the Locality--Capability balance under both preservation
choices.
These results support the interpretation that KLOD's advantage arises from the combination of bounded target optimization and target-excluded preservation, rather than from generic KL regularization alone.

\begin{table}[t]
\centering
\small
\begin{tabular}{lcccc}
\toprule
Method & Rel. & Gen. & Loc. & Cap. \\
\midrule
KLOD & 99.89 & 86.85 & 86.47 & 61.42 \\
w/o $L_{\mathrm{prefix}}$ & 99.97 & 88.67 & 81.26 & 60.92 \\
w/o $L_{\mathrm{nt}}$ & 99.57 & 87.99 & 82.09 & 61.68 \\
w/o $L_{\mathrm{prefix}}$, $L_{\mathrm{nt}}$ & 99.92 & 89.25 & 50.93 & 61.78 \\
w/o All & 99.93 & 89.21 & 51.24 & 62.10 \\
\bottomrule
\end{tabular}
\caption{
Ablation study on ZsRE with Llama3-8B-Instruct.
The row without both preservation terms uses only the bounded logit-odds objective.
The w/o All variant corresponds to LocFT-BF with the standard cross-entropy objective.
}
\label{tab:ablation_zsre}
\end{table}

\section{Additional Ablation on ZsRE}
\label{sec:ablation_zsre}

Table~\ref{tab:ablation_zsre} presents additional ablation results on ZsRE with Llama3-8B-Instruct.
All variants achieve high Reliability, indicating that the target edits remain effective under different objective variants.
Unlike the CounterFact setting, the bounded logit-odds objective alone does not provide a clear advantage over the standard cross-entropy counterpart on ZsRE.
Specifically, removing both preservation terms yields 50.93 Locality, which is comparable to the w/o All variant corresponding to LocFT-BF, which obtains 51.24 Locality.
Generalization and Capability are also nearly unchanged between these two variants.

Nevertheless, the KL-based preservation terms remain important for locality preservation.
Removing either $L_{\mathrm{prefix}}$ or $L_{\mathrm{nt}}$ decreases Locality from 86.47 to 81.26 and 82.09, respectively, while removing both preservation terms causes a much larger drop to 50.93.
This indicates that the two preservation terms are complementary on ZsRE: each term contributes moderately in isolation, but their combination is necessary to maintain high Locality.
Overall, the ZsRE ablation confirms that KLOD's locality improvement mainly comes from distribution-preserving regularization rather than from the bounded target objective alone.

\section{Post-hoc Decomposition of Locality Prediction Changes}
\label{app:locality_decomposition}

To better characterize locality degradation, we conduct a post-hoc
analysis of how locality predictions change after editing.
We decompose each locality prediction change into two mutually
exclusive categories: \emph{Target flip} ($F$) and
\emph{Other change} ($O$).

For each edit request $i$, let $T_i$ denote the set of content-token
IDs in its new target, excluding special and whitespace-only tokens.
At each teacher-forced locality-answer position $t$, let
$\hat{y}^{\mathrm{pre}}_{it}$ and
$\hat{y}^{\mathrm{post}}_{it}$ denote the pre- and post-edit
full-vocabulary top-1 predictions, respectively.

We first define a locality prediction change as
\begin{equation}
D_{it}
=
\mathbb{I}
\left[
\hat{y}^{\mathrm{post}}_{it}
\neq
\hat{y}^{\mathrm{pre}}_{it}
\right].
\label{eq:locality_change}
\end{equation}

A \emph{Target flip} ($F_{it}$) occurs when the prediction changes
from a non-target token to a token contained in the corresponding
edit target:
\begin{equation}
F_{it}
=
\mathbb{I}
\left[
\hat{y}^{\mathrm{pre}}_{it}\notin T_i
\;\land\;
\hat{y}^{\mathrm{post}}_{it}\in T_i
\right].
\label{eq:target_flip}
\end{equation}
Because the two membership conditions imply a change in the top-1
prediction, $D_{it}=1$ need not be stated separately.

All remaining locality prediction changes are defined as
\emph{Other change} ($O_{it}$):
\begin{equation}
O_{it}
=
D_{it}-F_{it}.
\label{eq:other_change}
\end{equation}
Therefore,
\begin{equation}
D_{it}=F_{it}+O_{it}
\label{eq:locality_token_decomposition}
\end{equation}
holds exactly at every evaluated token position.

We apply the same hierarchical aggregation as the official Locality
evaluation: token-level values are averaged within each locality
prompt, then at the edit-request level, and finally macro-averaged
over the 3,000 requests.
Let $F_D$ and $O_D$ denote the resulting dataset-level Target flip
and Other change scores for dataset $D$, respectively.
All quantities are recomputed through the same BF16 evaluation path
as the reported Locality scores.
Thus, before rounding,
\begin{equation}
\underbrace{100-\mathrm{Locality}_D}_{\mathrm{Damage}_D}
=
F_D + O_D.
\label{eq:locality_decomposition}
\end{equation}

\begin{table}[t]
\centering
\small
\setlength{\tabcolsep}{3.5pt}
\begin{tabular}{@{}l|rrrr@{}}
\toprule
Setting & Loc. & Damage & $F$ & $O$ \\
\midrule

\multicolumn{5}{@{}l}{\textit{CounterFact}} \\
\quad Bounded only
& 2.02 & 97.98 & 18.95 & 79.03 \\
\quad $+\,L_{\mathrm{nt}}$
& 41.30 & 58.70 & 13.57 & 45.13 \\
\quad Full KLOD
& 44.75 & 55.25 & 12.02 & 43.23 \\

\midrule

\multicolumn{5}{@{}l}{\textit{ZsRE}} \\
\quad Bounded only
& 50.93 & 49.07 & 0.094 & 48.974 \\
\quad $+\,L_{\mathrm{nt}}$
& 81.26 & 18.74 & 0.068 & 18.670 \\
\quad Full KLOD
& 86.47 & 13.53 & 0.050 & 13.477 \\

\bottomrule
\end{tabular}
\caption{
Post-hoc decomposition of locality prediction changes with
Llama3-8B-Instruct.
$F$ denotes flips to edit-target tokens, and $O$ denotes all other
prediction changes.
For ZsRE, $F$ and $O$ are reported to three decimal places because
target flips are rare.
}
\label{tab:locality_decomposition}
\end{table}

Table~\ref{tab:locality_decomposition} shows substantially different
failure patterns across datasets.
On CounterFact, $F$ represents a noticeable but non-dominant component
of locality damage.
With the bounded objective alone, $F$ accounts for 18.95 percentage
points of the 97.98-point Damage, while $O$ accounts for 79.03.
Adding $L_{\mathrm{nt}}$ reduces $F$ to 13.57 and $O$ to 45.13,
while full KLOD further reduces them to 12.02 and 43.23,
respectively.
Thus, preservation reduces both direct flips to edit-target tokens and
the remaining top-1 prediction changes.

The pattern is markedly different on ZsRE.
Although the bounded-only setting incurs 49.07 points of Damage,
$F$ accounts for only 0.094 points, whereas $O$ accounts for 48.974.
Likewise, $F$ remains extremely small after adding $L_{\mathrm{nt}}$
(0.068) and with full KLOD (0.050), even as Locality improves
substantially.
This indicates that locality degradation on ZsRE is not primarily
characterized by direct flips to edit-target tokens.

Overall, $F$ captures one observable form of locality failure, but its
prevalence differs substantially across datasets.
The reduction in $O$ further shows that KLOD's locality improvement is
not limited to preventing edited target tokens from directly replacing
unrelated predictions, consistent with the broader distributional
preservation behavior observed in our KL analyses.

\section{Free-Running Generation Evaluation}
\label{app:free_running}

The primary Reliability and Generalization metrics in our experiments
follow the standard teacher-forcing protocol used in knowledge editing.
To examine whether the observed behavior also holds under actual
generation, we additionally perform a free-running evaluation after
3k CounterFact edits with Llama3-8B-Instruct.

We sample 500 evaluation cases from the 3k CounterFact set using
random seed 42 and use the same cases for all methods.
For each rewrite and rephrase prompt, the edited model generates the
answer autoregressively without access to the gold target tokens.
We use greedy decoding with a maximum of 32 generated tokens.
The generated answer is truncated at the first period or newline and
normalized by lowercasing, removing punctuation and articles, and
collapsing whitespace.
Reliability and Generalization are then measured using normalized
exact match for the rewrite and rephrase prompts, respectively.

\begin{table}[t]
\centering
\small
\begin{tabular}{lccc}
\toprule
Method & Rel. EM & Gen. EM & Cap. \\
\midrule
pre-edit
& 0.00 & 0.20 & 62.11 \\
AlphaEdit
& 66.60 & 52.20 & 25.46 \\
LocFT-BF
& \textbf{99.80} & 53.00 & 41.92 \\
OVERTONE
& 99.60 & 56.40 & 42.51 \\
KLOD ($\alpha=0.85$)
& 99.60 & 48.00 & \textbf{60.77} \\
KLOD ($\alpha=1.0$)
& 99.60 & \textbf{61.60} & 57.65 \\
\bottomrule
\end{tabular}
\caption{
Free-running generation results on 500 CounterFact cases after 3k
edits with Llama3-8B-Instruct.
Rel./Gen. EM denote normalized exact match on rewrite/rephrase prompts;
bold marks the best edited-model result in each column.
}
\label{tab:free_running}
\end{table}

Table~\ref{tab:free_running} shows that the main conclusions are not
specific to teacher-forced token prediction.
At the primary locality-oriented setting, KLOD with
$\alpha=0.85$ achieves 99.60 Reliability EM and preserves Capability
at 60.77, close to the pre-edit score of 62.11.
In contrast, LocFT-BF and OVERTONE obtain Capability scores of 41.92
and 42.51, respectively.

The generation results also reproduce the controllable
Generalization trade-off observed in the teacher-forced evaluation.
Increasing the target threshold to $\alpha=1.0$ raises
Generalization EM from 48.00 to 61.60 while maintaining 99.60
Reliability EM and 57.65 Capability.
This exceeds the generation-based Generalization of LocFT-BF
(53.00) and OVERTONE (56.40), consistent with the
$\alpha$-controlled operating curve observed in
Section~\ref{sec:alpha_sweep}.

We do not report free-running Locality in this analysis because the
main Locality metric is defined by agreement between the pre-edit and
post-edit model predictions on unrelated prompts.
A ground-truth exact-match score on locality answers would measure a
different quantity and would therefore not be directly comparable to
the Locality metric reported in the main experiments.

\begin{table}[t]
\centering
\small
\begin{tabular}{lcccc}
\toprule
Method & Rel. & Gen. & Loc. & Cap. \\
\midrule
LocFT-BF & 99.93 & \textbf{96.21} & 35.22 & \textbf{61.83} \\
OVERTONE & 99.90 & 95.98 & 34.98 & 61.62 \\
KLOD     & 99.75 & 93.11 & \textbf{56.30} & 61.30 \\
\bottomrule
\end{tabular}
\caption{
Results after 3k edits on WikiBigEdit with Llama3-8B-Instruct.
}
\label{tab:wikibigedit}
\end{table}

\section{Additional Evaluation on WikiBigEdit}
\label{app:wikibigedit}

To evaluate whether KLOD's behavior extends beyond CounterFact and ZsRE, we additionally conduct a 3k-edit evaluation on WikiBigEdit~\citep{WikiBigEdit} with Llama3-8B-Instruct.

In Table~\ref{tab:wikibigedit}, KLOD achieves 56.30 Locality, improving over LocFT-BF and OVERTONE
by 21.08 and 21.32 percentage points, respectively.
This improvement comes with a modest reduction in Generalization,
while Reliability remains above 99\% and Capability remains comparable
to the two fine-tuning baselines.
These results provide additional evidence that KLOD's locality
preservation benefit extends to WikiBigEdit.

\begin{table}[t]
\centering
\small
\begin{tabular}{ccccc}
\toprule
$\lambda_{\mathrm{nt}}$ & Rel. & Gen. & Loc. & Cap. \\
\midrule
0.2 & 99.70 & 48.23 & 35.33 & 60.98 \\
0.4 & 99.63 & 47.43 & 40.37 & 60.78 \\
0.6 & 99.70 & 47.37 & 44.75 & 60.77 \\
0.8 & 99.63 & 45.07 & 47.42 & 61.07 \\
\bottomrule
\end{tabular}
\caption{
Sweep over $\lambda_{\mathrm{nt}}$ with $\lambda_{\mathrm{prefix}}$ fixed at $1.2$ on CounterFact with Llama3-8B-Instruct.
}
\label{tab:lambda_nt_sweep}
\end{table}

\begin{table}[t]
\centering
\small
\begin{tabular}{ccccc}
\toprule
$\lambda_{\mathrm{prefix}}$ & Rel. & Gen. & Loc. & Cap. \\
\midrule
0.5 & 99.77 & 48.37 & 44.25 & 60.88 \\
0.9 & 99.77 & 47.27 & 44.75 & 60.85 \\
1.2 & 99.70 & 47.37 & 44.75 & 60.77 \\
1.5 & 99.73 & 46.20 & 44.97 & 60.68 \\
\bottomrule
\end{tabular}
\caption{
Sweep over $\lambda_{\mathrm{prefix}}$ at the primary operating
point, with $\lambda_{\mathrm{nt}}=0.6$ and $\alpha=0.85$, on
CounterFact with Llama3-8B-Instruct.
}
\label{tab:lambda_prefix_sweep}
\end{table}

\section{Lambda Sweep}
\label{app:lambda_sweep}

This section analyzes how the two preservation loss weights in KLOD, $\lambda_{\mathrm{nt}}$ and $\lambda_{\mathrm{prefix}}$, affect editing performance.
All experiments are conducted on CounterFact with 3k sequential edits using Llama3-8B-Instruct, with the target probability threshold fixed at $\alpha=0.85$.

Table~\ref{tab:lambda_nt_sweep} shows the results of varying $\lambda_{\mathrm{nt}}$ while fixing $\lambda_{\mathrm{prefix}}=1.2$.
As $\lambda_{\mathrm{nt}}$ increases, Locality improves from 35.33 to 47.42, while Generalization decreases from 48.23 to 45.07.
Reliability and Capability remain nearly stable, indicating that stronger non-target distribution preservation improves locality with a moderate trade-off in rephrase generalization.

Table~\ref{tab:lambda_prefix_sweep} varies $\lambda_{\mathrm{prefix}}$ while fixing $\lambda_{\mathrm{nt}}=0.6$, corresponding to the primary KLOD operating point.
Increasing $\lambda_{\mathrm{prefix}}$ produces only a small Locality increase and a modest Generalization decrease, while Reliability and Capability remain nearly unchanged.
Together, the two sweeps show that $L_{\mathrm{nt}}$ is the primary preservation component on CounterFact, whereas $L_{\mathrm{prefix}}$ acts as a gentler complementary regularizer whose marginal contribution is larger on ZsRE, as shown in Appendix~\ref{sec:ablation_zsre}.


\end{document}